\documentclass[11pt]{article}
\usepackage{algorithm}  
\usepackage{algpseudocode}
\usepackage[margin=1in]{geometry}
\usepackage[numbers,sort&compress]{natbib}
\usepackage{bm}

\usepackage[utf8]{inputenc} 
\usepackage[T1]{fontenc}    
\usepackage{hyperref}       
\usepackage{url}            
\usepackage{booktabs}       
\usepackage{amsfonts}  

\usepackage{nicefrac}       
\usepackage{microtype}  
\usepackage{graphicx}
\usepackage{xcolor}         
\usepackage{enumitem}
\usepackage{multirow}
\usepackage{mathtools}
\usepackage{amssymb}
\usepackage{colortbl}
\usepackage{graphicx}
\usepackage{url}
\usepackage{pgfplots}  
\usepackage{caption}      
\usepackage{subcaption} 
\usepackage{amsmath}
\usepackage{booktabs}
\usepackage{wrapfig}
\usepackage{float}
\usepackage{enumitem}

\newtheorem{lemma}{Lemma}
\newtheorem{Definition}{Definition}
\newtheorem{Assumption}{Assumption}
\newtheorem{Proposition}{Proposition}
\newtheorem{Theorem}{Theorem}
\newtheorem{Remark}{Remark}
\newtheorem{Main Theorem}{Main Theorem}

\newcommand{\gr}{\nabla}
\makeatletter
\renewcommand*{\@fnsymbol}[1]{\ensuremath{\ifcase#1\or \dagger\or \ddagger
\else\@ctrerr\fi}}
\makeatother

\title{Robust Decentralized Learning with Local Updates
and Gradient Tracking}
\author{
Sajjad Ghiasvand~\thanks{Electrical and Computer Engineering Department, UC Santa Barbara, Santa Barbara, CA, USA}
\and Amirhossein Reisizadeh~\thanks{Laboratory for
Information and Decision Systems, Massachusetts Institute of Technology, Cambridge, MA, USA}
\and Mahnoosh Alizadeh~\footnotemark[1]
\and Ramtin Pedarsani~\footnotemark[1]
}

\begin{document}

\maketitle
\begin{abstract}
\normalsize
As distributed learning applications such as Federated Learning, the Internet of Things (IoT), and Edge Computing grow, it is critical to address the shortcomings of such technologies from a theoretical perspective. As an abstraction, we consider decentralized learning over a network of communicating clients or nodes and tackle two major challenges: \emph{data heterogeneity} and \emph{adversarial robustness}. We propose a decentralized minimax optimization method that employs two important modules: local updates and gradient tracking. Minimax optimization is the key tool to enable adversarial training for ensuring robustness. Having local updates is essential in Federated Learning (FL) applications to mitigate the communication bottleneck, and utilizing gradient tracking is essential to proving convergence in the case of data heterogeneity. We analyze the performance of the proposed algorithm, Dec-FedTrack, in the case of nonconvex-strongly-concave minimax optimization, and prove that it converges a stationary point.  We also conduct numerical experiments to support our theoretical findings.
\end{abstract}

\section{Introduction}
Learning from distributed data is at the core of modern and successful technologies such as Internet of Things (IoT), Edge Computing, fleet learning, etc., where massive amounts of data are generated across dispersed users. Depending on the application, there are two main architectures for the learning paradigm: (i) A \emph{distributed} setting with a central parameter server or master nodes that are responsible for aggregating the model and is able to communicate to all the computing nodes or workers; (ii) A \emph{decentralized} setting for which there is no central coordinating node, and all the nodes communicate to their neighbors through a connected communicating graph. In this work, we focus on the latter. 

Federated learning (FL) is a novel and promising distributed learning paradigm mostly employed using the master-worker architecture that aims to find accurate models across distributed nodes \cite{konevcny2016federated,kairouz2021advances}. The main premise of FL framework is user data privacy, that is, locally stored data on each entity remains local during the training procedure, which is in contrast to traditional distributed learning paradigms. In the peer-to-peer or decentralized implementation of FL methods which is the focus of this work, distributed nodes update model parameters locally using local optimization modules such as Stochastic Gradient Descent (SGD) and exchange information with their neighboring nodes to reach consensus. In Federated Learning, due to privacy and communication constraints, each communication round consists of \emph{multiple local updates} before each node aggregates the neighboring updates.


While FL enables us to efficiently train a model, an important challenge is to ensure the robustness of the learned model to possible noisy/adversarial perturbations~\cite{nabavirazavi2024enhancing}. The problem becomes more critical in FL since due to its distributed nature, it is more vulnerable to the presence of adversarial nodes and adversarial attacks~\cite{nabavirazavi2024impact,nabavirazavi2023model}. Adversarial training based on minimax optimization is the key tool to robustify the learned model in machine learning applications~\cite{saberi2025detecting}. Thus, it is critical to develop decentralized minimax optimization algorithms that are also communication-efficient, i.e. optimization methods that employ local updates suitable for a federated setting. Other applications of federated minimax optimization include using optimal transport to develop personalized FL \cite{farnia2022optimal} and robustness against distributed shifts \cite{reisizadeh2020robust}. Another major challenge in decentralized learning methods is data heterogeneity. Data heterogeneity refers to the fact that the data distributions across distributed nodes are statistically heterogeneous (or non-iid). In this work, we employ the \emph{gradient tracking} (GT) technique that guarantees convergence of the algorithm in the presence of data heterogeneity. 

\noindent\textbf{Contributions.}
We propose the  Dec-FedTrack algorithm which is a decentralized minimax optimization method over a network of $n$ communicating nodes with two modules of local updates and gradient tracking, and analyze its communication complexity and convergence rate for the case of nonconvex-strongly-concave (NC-SC) minimax optimization. We show that Dec-FedTrack achieves the $O\left(\kappa^5 n^{-1}\epsilon^{-4} \right)$ stochastic first-order oracle (SFO) complexity and the $O\left(\kappa^3\epsilon^{-2} \right)$ communication complexity, where the condition number is defined by $ \kappa \triangleq \ell/\mu $. This is the first federated minimax optimization algorithm that incorporates GT in a decentralized setting.
Moreover, we conduct several numerical experiments that demonstrate the communication efficiency and adversarial robustness of Dec-FedTrack over baselines.

\section{Related Work}\label{sec:relwork}
\subsection{Federated Learning with Heterogeneous Data}
One of the most challenging aspects of federated learning is data heterogeneity, where training data is not identically and independently distributed across clients (non-i.i.d.). Under such conditions, local models of clients may drift away from the global model optimum, slowing down convergence \cite{hsieh2020non,li2020federated}. Several studies have attempted to tackle this issue in federated learning \cite{zhou2023fedpage,rodio2023federated,wang2023bose,liu2024federated}. However, these studies are typically not decentralized, their results are often limited to (strongly) convex objective functions, or they make restrictive assumptions about the gradients of objective functions. In this context, gradient tracking (GT) algorithms have been proposed to address these challenges \cite{koloskova2021improved,pu2021distributed,zhang2019decentralized,ebrahimi2024distributed}. Particularly, in this paper, we also leverage the GT technique to mitigate the data heterogeneity problem.

\subsection{Decentralized Minimization}
  Many works have examined minimization problems within a decentralized setting~\cite{chen2021communication,hendrikx2021optimal,kovalev2020optimal,li2020communication,li2022destress, li2022variance,sun2020improving,uribe2020dual,wang2021distributed,xin2022fast,liu2023yoga}. Works such as K-GT~\cite{liu2023decentralized}, LU-GT~\cite{nguyen2022performance} and \cite{koloskova2021improved,haddadpour2021federated,berahas2023balancing,pu2021distributed} have introduced decentralized algorithms incorporating local updates and GT, although they are tailored for minimization rather than minimax optimization.
\subsection{Centralized Minimax Optimization}
Centralized minimax optimization has become increasingly significant, particularly with the rise of machine learning applications like GANs~\cite{goodfellow2014generative} and adversarial training of neural networks. This optimization paradigm tackles the challenges posed by nonconvex-concave and nonconvex-nonconcave problems, drawing attention due to its relevance in various domains. For NC-SC problems, several works have utilized momentum or variance reduction techniques to achieve the SFO complexity of $O\left(\kappa^3\epsilon^{-3} \right)$~\cite{lin2020gradient,qiu2020single,luo2020stochastic,zhang2021complexity}. 
\subsection{Decentralized Minimax Optimization}
Numerous studies have explored decentralized minimax optimization for (strongly) convex-concave~\cite{koppel2015saddle, mateos2015distributed,rogozin2024decentralized,beznosikov2021distributed,beznosikov2021near}, nonconvex-strongly-concave~\cite{wu2023decentralized, liu2023precision, xu2023decentralized, gao2022decentralized, mancino2023variance, chen2022simple, xian2021faster,zhang2023jointly,tsaknakis2020decentralized}, and nonconvex-nonconcave~\cite{liu2020decentralized}, objective functions. DPOSG~\cite{liu2020decentralized} has the assumption of identical distributions, and most of the mentioned works on nonconvex-strongly-concave minimax optimization have a very high gradient complexity. The closest ones to our setting and results are DM-HSGD~\cite{xian2021faster}, DREAM~\cite{chen2022simple}, and black~\cite{zhang2023jointly}. These studies explore decentralized minimax optimization using gradient tracking and variance reduction techniques. DM-HSGD employs the variance reduction technique of STORM~\cite{cutkosky2019momentum}, whereas DREAM and black utilize the variance reduction technique of SPIDER~\cite{fang2018spider}. However, clients in these algorithms do not perform multiple local updates between communication rounds, making them unsuitable for federated learning scenarios.
\subsection{Distributed/Federated Minimax Learning}
Several works have studied minimax optimization in the federated learning setting across various function types: (strongly) convex-concave~\cite{reisizadeh2020robust, hou2021efficient,liao2021local,sun2022communication} and nonconvex-strongly-concave/nonconvex-PL/nonconvex-one-point-concave~\cite{deng2020distributionally, deng2021local, xie2021federated,sharma2022federated, sharma2023federated}. FedGDA-GT~\cite{sun2022communication} has delved into federated minimax learning with both local updates and GT, but it is not decentralized and assumes strongly-convex-strongly-concave objective functions. Momentum Local SGDA~\cite{sharma2022federated}, SAGDA~\cite{yang2022sagda}, and De-Norm-SGDA~\cite{sharma2023federated} explore federated minimax optimization with local updates but lacks decentralization and does not incorporate GT.

 We summarize the comparison of related algorithms with Dec-FedTrack in Table~\ref{comparison}.

\begin{table*}[t]
\centering
\caption{Comparison of Dec-FedTrack with related algorithms for minimax and minimization optimization. Criteria in this comparison are: SFO complexity; number of communications; type of centralization; type of function class; if the algorithm is stochastic; and if the algorithm has local update (LU), heterogeneity robustness (HR), and adversarial robustness (AR).}
\label{comparison}
\renewcommand{\arraystretch}{1.5} 
\resizebox{\textwidth}{!}{%
\begin{tabular}{ccccccccc}
\hline
\text{ Name } & \text{ SFO } & \text{ Comm. Round } & \text{ Decentralized } & \text{ Objective } & \text{ LU } & \text{ HR } & \text{ AR }\\
\hline
\text{ MLSGDA~\cite{sharma2022federated} }  & $O\left(\frac{\kappa^4}{n\epsilon^{4}} \right)$ & $O\left(\frac{\kappa^3}{\epsilon^{3}}\right)$ & \texttimes & \text{ NC-SC } & \checkmark & \texttimes & \checkmark \\
\text{ SAGDA~\cite{yang2022sagda} }  & $O\left(\frac{\kappa^4}{n\epsilon^{4}} \right)$ & $O\left(\frac{\kappa^2}{\epsilon^{2}}\right)$ & \texttimes & \text{ NC-SC } & \checkmark & \texttimes & \checkmark \\
\text{ Fed-Norm-SGDA~\cite{sharma2023federated} }  & $O\left(\frac{\kappa^4}{n\epsilon^{4}} \right)$ & $O\left(\frac{\kappa^2}{\epsilon^{2}}\right)$ & \texttimes & \text{ NC-SC } & \checkmark & \texttimes & \checkmark \\
\text{ DM-HSGD~\cite{xian2021faster} } & $O\left(\frac{\kappa^3}{n\epsilon^{3}} \right)$ & $O\left(\frac{\kappa^3}{\epsilon^{3}} \right)$ & \checkmark & \text{ NC-SC } & \texttimes & \checkmark & \checkmark \\
\text{ DREAM~\cite{chen2022simple} } & $O\left(\frac{\kappa^3}{n\epsilon^{3}}\right)$ & $O\left(\frac{\kappa^2}{\epsilon^{2}}\right)$ & \checkmark & \text{ NC-SC } & \texttimes & \checkmark & \checkmark \\
\text{ black~\cite{zhang2023jointly} } & $O\left(\frac{\kappa^4}{n\epsilon^{3}}\right)$ & $O\left(\frac{\kappa^3}{\epsilon^{2}}\right)$ & \checkmark & \text{ NC-SC } & \texttimes & \checkmark & \checkmark \\
\text{ K-GT~\cite{liu2023decentralized} } & $O\left(\frac{1}{n\epsilon^{4}} \right)$ & $O\left(\frac{1}{\epsilon^{2}} \right)$ & \checkmark & \text{ NC } & \checkmark & \checkmark & \texttimes \\
\text{ Dec-FedTrack (Ours) } & $O\left(\frac{\kappa^5}{n\epsilon^{4}} \right)$ & $O\left(\frac{\kappa^3}{\epsilon^{2}} \right)$ & \checkmark & \text{ NC-SC } & \checkmark & \checkmark & \checkmark \\ [1mm]
\hline
\end{tabular}%
}
\end{table*}

\section{Problem Setup}
We consider a connected network of $ n $ clients with $ \mathcal{V} =  [n] := \{1,\ldots n\} $ and $ \mathcal{E} \subseteq \mathcal{V}\times\mathcal{V} $ as the set of nodes and edges, respectively.
This network collaboratively seeks to solve the following minimax optimization problem: 
\begin{align}\label{eq: main minimax}
    \min _{\mathbf{x} \in \mathbb{R}^{d}} \max_{\mathbf{y} \in \mathbb{R}^q} f(\mathbf{x}, \mathbf{y})
    =
    \frac{1}{n} \sum_{i=1}^n f_i(\mathbf{x}, \mathbf{y}),
\end{align}
where $f_i(\mathbf{x}, \mathbf{y}) = \mathbb{E}[F_i(\mathbf{x}, \mathbf{y} ; \xi^{(i)})]$ denotes the local function associated with node $i \in \mathcal{V}$. Here, the expectation is with respect to $\xi^{(i)} \sim \mathcal{D}_i$ and $ \mathcal{D}_i $ denotes the local distribution for node $i$. 
In our decentralized setting, clients communicate with each other along the edges $ e \in \mathcal{E} $, that is, each node is allowed to communicate with its neighboring nodes.

\subsection{Motivating example: Federated adversarial training}

Consider a network of clients that wish to train a common model $\mathbf{x}$ that is robust to adversarial perturbation $\mathbf{y}$. In this model, the adversary can attack the network by adding a common perturbation to \emph{all} the samples of every node, i.e. \emph{universal perturbation} \cite{moosavi2017universal,oskouie2023interpretation}. This model corresponds to an adversarial cost function $f_i(\mathbf{x}, \mathbf{y})$ for each node $i$ and results in a minimax problem shown in \eqref{eq: main minimax} that should be solved over the connected network. One should add that in adversarial machine learning, the adversary is restricted to a bounded noise power; therefore, in this case, the minimax problem \eqref{eq: main minimax} will have a constraint $\| \mathbf{y} \| \leq \delta$.

\subsection{Convergence measure}

In this paper, we focus on a particular setting where each local function $f_i(\mathbf{x}, \mathbf{y})$ is nonconvex in $\mathbf{x}$ and strongly concave in $\mathbf{y}$ which is well-studied in the minimax optimization literature \cite{lin2020near}. This assumption allows us to define the \emph{primal} function of \eqref{eq: main minimax} for every $\mathbf{x}$ as $\Phi(\mathbf{x}) \coloneqq \max _{\mathbf{y} \in \mathbb{R}^q} f(\mathbf{x}, \mathbf{y})$.
Solving the minimax problem \eqref{eq: main minimax} is equivalent to minimizing the primal function, i.e., $\min_{\mathbf{x} \in \mathbb{R}^d} \Phi(\mathbf{x})$ which is nonconvex. A well-established convergence measure for such minimization problems is to find a \emph{stationary point} $\hat{\mathbf{x}}$ of $\Phi$, that is a point for which $\| \gr \Phi(\hat{\mathbf{x}}) \| \leq \epsilon$.

\subsection{Notation}
We represent vectors using bold small letters and matrices using bold capital letters. The vector $\mathbf{x}_i^{(t)+k}$ denotes a variable on node $i$ at local step $k$ and communication round $t$, as will be explained in Section \ref{Algorithm}. The average of vectors $ \mathbf{x}_i $ is defined as $ \bar{\mathbf{x}} = \frac{1}{n}\sum_i \mathbf{x}_i $. We denote a matrix whose columns are the collection of $n$ vectors, each belonging to a client, as $ \mathbf{X}\in \mathbb{R}^{d\times n} $, i.e., $ \mathbf{X} = [\mathbf{x}_1, \cdots, \mathbf{x}_n]$. Additionally, we use $ \bar{\mathbf{X}} $ to represent a matrix whose columns are equal to $ \bar{\mathbf{x}}$, and it can be written in a more useful way as
\begin{align}
    \bar{\mathbf{X}}=[\bar{\mathbf{x}}, \ldots, \bar{\mathbf{x}}]=\frac{1}{n} \mathbf{X} \mathbf{1}_n \mathbf{1}_n^T = \mathbf{X}\mathbf{J} \in \mathbb{R}^{d \times n},\nonumber
\end{align}
where $ \mathbf{J} = \frac{1}{n}\mathbf{1}_n \mathbf{1}_n^T $. We also use the below notation for convenience throughout the paper:
\begin{gather}
\nabla F(\mathbf{X}, \mathbf{Y} ; \xi)  =\left[\nabla F_1\left(\mathbf{x}_1, \mathbf{y}_1 ; \xi_1\right), \ldots, \nabla F_n\left(\mathbf{x}_n, \mathbf{y}_n ; \xi_n\right)\right], \nonumber\\
  \nabla f(\mathbf{X}, \mathbf{Y})  =\mathbb{E}_{\left(\xi_1, \ldots, \xi_n\right)} \nabla F(\mathbf{X}, \mathbf{Y} ; \xi)=\left[\nabla f_1\left(\mathbf{x}_1, \mathbf{y}_1\right), \ldots, \nabla f_n\left(\mathbf{x}_n, \mathbf{y}_n\right)\right] \in \mathbb{R}^{d \times n}.   \nonumber
\end{gather}

We denote the batch sizes for variables $ \mathbf{x} $ and $ \mathbf{y} $ as $ b_x $ and $ b_y $, respectively.

\section{Proposed Algorithm}

In this section, we describe our proposed method to solve the minimax problem \eqref{eq: main minimax} over a connected network of $n$ nodes. Our method, namely Dec-FedTrack, comprises of two main modules: \emph{local updates} and \emph{gradient tracking} which we elaborate on in the following.

Dec-FedTrack (shown in Algorithm \ref{al.1}) consists of a number of communication rounds, $T$, where in each round, every node performs $K$ local updates on its variables.
In particular, in the $k$th iteration of round $t$, each node computes unbiased stochastic gradients and updates its local min and max variables $\mathbf{x}_i$ and $\mathbf{y}_i$ using the so-called \emph{correction terms} (Lines 4 and 5). Next, each node obtains tracking variables
\begin{align}
    \mathbf{z}_i^{(t)}&=\frac{1}{K \eta_c}(\mathbf{x}_i^{(t)}-\mathbf{x}_i^{(t)+K}),\nonumber \\
    \mathbf{r}_i^{(t)}&=\frac{1}{K \eta_d}(\mathbf{y}_i^{(t)+K}-\mathbf{y}_i^{(t)}),\nonumber
\end{align}
and sends variable $ \{\mathbf{z}_i^{(t)}, \mathbf{r}_i^{(t)}, \mathbf{x}_i^{(t)}, \mathbf{y}_i^{(t)} \} $ to its neighboring nodes. After aggregating these variables from the neighbors, node $i$ updates its correction terms and model variables using gradient tracking \cite{liu2023decentralized} as follows:
\begin{align}
    \mathbf{c}_i^{(t+1)}&=\mathbf{c}_i^{(t)}-\mathbf{z}_i^{(t)}+\sum_j w_{i j} \mathbf{z}_j^{(t)},\nonumber\\
    \mathbf{d}_i^{(t+1)}&=\mathbf{d}_i^{(t)}-\mathbf{r}_i^{(t)}+\sum_j w_{i j} \mathbf{r}_j^{(t)},\nonumber\\
    \mathbf{x}_i^{(t+1)}&=\sum_j w_{i j}\left(\mathbf{x}_j^{(t)}-K \eta_x \mathbf{z}_j^{(t)}\right),\nonumber\\
    \mathbf{y}_i^{(t+1)}&=\sum_j w_{i j}\left(\mathbf{y}_j^{(t)}+K \eta_y \mathbf{r}_j^{(t)}\right),\nonumber
\end{align}
where $ \eta_x := \eta_s \eta_c  $ and $ \eta_y := \eta_r \eta_d $ denote the global step sizes. The proposed Dec-FedTrack algorithm is described in Algorithm \ref{al.1} using matrix notations.

\begin{algorithm}[t!]
	\caption{Dec-FedTrack} \label{al.1}
 \noindent\textbf{Initialize:} $\forall i, j \in [n], \mathbf{x}_i^{(0)}=\mathbf{x}_j^{(0)}, \mathbf{y}_i^{(0)}=\mathbf{y}_j^{(0)}$; $ \mathbf{c}_i^{(0)} $ and $ \mathbf{d}_i^{(0)} $ according to Lemma \ref{lem.3}.
	\begin{algorithmic}[1]

			\For {\textbf{communication:} $ t\leftarrow0 $ to $ T - 1 $}
			        \For {node $ i \in [n] $ parallel}
				\For {\textbf{local step:} $ k\leftarrow0 $ to  $ K - 1 $}
				\State Update min variables
				    \begin{align}
				        \mathbf{X}^{(t)+k+1} =
                            \mathbf{X}^{(t)+k}-\eta_c(\nabla_x F(\mathbf{X}^{(t)+k}, \mathbf{Y}^{(t)+k};  \xi^{(t)+k})+\mathbf{C}^{(t)})\nonumber
				    \end{align}
				\State Update max variables
                        \begin{align}
				        \mathbf{Y}^{(t)+k+1} =
                            \mathbf{Y}^{(t)+k}+\eta_d(\nabla_y F(\mathbf{X}^{(t)+k}, \mathbf{Y}^{(t)+k};  \xi^{(t)+k})+\mathbf{D}^{(t)})\nonumber
				    \end{align}
				\EndFor
				\State $ \mathbf{Z}^{(t)} = \frac{1}{K\eta_c}\left( \mathbf{X}^{(t)} - \mathbf{X}^{(t) + K} \right)   $              
                    \State$ \mathbf{R}^{(t)} = \frac{1}{K\eta_d}\left( \mathbf{Y}^{(t) + K} - \mathbf{Y}^{(t)} \right) $
				\State $ \mathbf{C}^{(t+1)} = \mathbf{C}^{(t)} - \mathbf{Z}^{(t)} + \mathbf{Z}^{(t)}\mathbf{W}$
                    \State $\mathbf{D}^{(t+1)} = \mathbf{D}^{(t)} - \mathbf{R}^{(t)} + \mathbf{R}^{(t)}\mathbf{W}  $
				\State  $ \mathbf{X}^{(t+1)} = \left( \mathbf{X}^{(t)} - K\eta_x \mathbf{Z}^{(t)}  \right)\mathbf{W} $ 
                    \State $\mathbf{Y}^{(t+1)} = \left( \mathbf{Y}^{(t)} + K\eta_y \mathbf{R}^{(t)}  \right)\mathbf{W} $
			\EndFor
			
		\EndFor
	\end{algorithmic} 
\end{algorithm}
Next, we comment on the necessity of using GT in our proposed algorithm. Given that clients' distributions are non-iid, to prove convergence one needs to establish an upper bound on the local gradients. While bounding assumptions can be directly imposed on local gradients, such as Assumption 3b in~\cite{koloskova2020unified}, in many distributed learning settings that are unconstrained, assuming the existence of such bounds can be restrictive. The gradient tracking algorithm~\cite{pu2021distributed} addresses this challenge by incorporating a correction term into gradients at each node. In fact, the correction term aims to bring the tracking variable for each client close to the tracking variable of its neighbors, preventing client-drift. 
The matrix format of the correction term in GT is as follows:
\begin{gather}
\mathbf{X}^{(t+1)}=\left(\mathbf{X}^{(t)}-\eta \mathbf{Z}^{(t)}\right) \mathbf{W}\nonumber \\ 
\mathbf{Z}^{(t+1)}=\nabla F\left(\mathbf{X}^{(t+1)} ; \xi^{(t+1)}\right) +\underbrace{\mathbf{Z}^{(t)} \mathbf{W}-\nabla F\left(\mathbf{X}^{(t)} ; \xi^{(t)}\right)}_{\text{correction term}}.\nonumber
\end{gather}

\section{Convergence Analysis}
In this section, we provide rigorous convergence analysis for the proposed Dec-FedTrack algorithm solving \eqref{eq: main minimax}.
We first present the following preliminary definitions for functions with one variable:

\begin{Definition}\label{def2} A function $ f$ is called $L$-Lipschitz if for any $\mathbf{x}$ and $\mathbf{x}'$, we have $ \left\|f(\mathbf{x})-f\left(\mathbf{x}^{\prime}\right)\right\| \leq L\left\|\mathbf{x}-\mathbf{x}^{\prime}\right\|$.
\end{Definition}
\begin{Definition}\label{def3} A function $ f $ is called $\ell$-smooth if it is diﬀerentiable and for any $\mathbf{x}$ and $\mathbf{x}'$, we have $\left\|\nabla f(\mathbf{x})-\nabla f\left(\mathbf{x}^{\prime}\right)\right\| \leq \ell\left\|\mathbf{x}-\mathbf{x}^{\prime}\right\|$.
\end{Definition}

Let us proceed with a few assumptions. 

As explained before, in our decentralized setting, agents communicate with each other along the edges of a fixed communication graph connecting $n$ nodes. Moreover, each edge of the graph is associated with a positive mixing weight and we denote the mixing matrix by $ \mathbf{W}\in \mathbb{R}^{n\times n} $.
\begin{Assumption}\label{assm: W}
The mixing matrix $ \mathbf{W}$ has the following properties: (i) Every element of $ \mathbf{W} $ is non-negative, and $ W_{i,j} = 0 $ if and only if $ i $ and $ j $ are not connected, (ii) $ \mathbf{W}\mathbf{1} = \mathbf{W}^{\top}\mathbf{1} = 1 $, (iii) there exists a constant $ 0\leq p \leq 1 $ such that 
\begin{align}
    \|\mathbf{X W}-\bar{\mathbf{X}}\|_F^2 \leq(1-p)\|\mathbf{X}-\bar{\mathbf{X}}\|_F^2, \forall \mathbf{X} \in \mathbb{R}^{d \times n}.\nonumber
\end{align}
\end{Assumption}
The mixing rate illustrates the degree of connectivity within the network. A higher $p$ signifies a more interconnected communication graph.  When $p=1$, $\mathbf{W}=\frac{1}{n} \mathbf{1}\mathbf{1}^T$, suggesting full connectivity in the graph, while $p=0$ yields $\mathbf{W}=\mathbf{I}_n$, indicating a disconnected graph~\cite{liu2023decentralized}.

\begin{Assumption} \label{assm: smooth}
    We assume that each local objective function $f_i$ is $\ell$-smooth, that is, for all $\mathbf{x},\mathbf{x}',\mathbf{y},\mathbf{y}'$
    \begin{align}
 \|\nabla f_i (\mathbf{x}, \mathbf{y} )-\nabla f_i(\mathbf{x}^{\prime}, \mathbf{y}^{\prime} )\|^2 \leq \ell^2(\|\mathbf{x}-\mathbf{x}^{\prime}\|^2+\|\mathbf{y}-\mathbf{y}^{\prime}\|^2).\nonumber
\end{align}
We also assume that each $f_i(\mathbf{x},\cdot)$ is $\mu$-strongly concave with respect to its second argument. We denote the condition number by $\kappa := {\ell}/{\mu} $.
\end{Assumption}
The above assumption implies that the objective function $f$ in \eqref{eq: main minimax} is $\ell$-smooth and strongly concave with respect to its second argument.

\begin{Assumption}\label{assm: stch gr}
We assume that the stochastic gradients are unbiased and variance-bounded, that is,
\begin{gather}
    \mathbb{E} \left[\nabla F_i(\mathbf{x}, \mathbf{y}; \xi_i)\right ] = \nabla f_i(\mathbf{x}, \mathbf{y}),\nonumber\\
    \mathbb{E}\|\nabla F_i(\mathbf{x}, \mathbf{y}; \xi_i)\ - \nabla f_i(\mathbf{x}, \mathbf{y})\|^2 \leq \sigma^2.\nonumber
\end{gather}
\end{Assumption}

\begin{Assumption}\label{assm: lower bound}
The function $ \Phi(\cdot) $ is lower bounded, that is $\inf _\mathbf{x} \Phi(\mathbf{x})=\Phi^*>-\infty.$
\end{Assumption}

Next, we provide the main result of the paper.

\begin{Theorem}\label{Theorem}
Suppose  Assumptions \ref{assm: W}-\ref{assm: lower bound} hold and consider the iterates of Dec-FedTrack in Algorithm \ref{al.1} with step-sizes  $\eta_d=$ $\Theta\left(\frac{p}{\kappa K \ell}\right), \eta_c=\Theta\left(\frac{\eta_d}{\kappa^2}\right)$, and $\eta_s=\eta_r=\Theta(p)$. Then, after $T$ communication rounds each with $K$ local updates, there exists an iterate $0 \leq t \leq T$ such that $\mathbb{E}\|\nabla \Phi(\bar{\mathbf{x}}^{(t)})\|^2 \leq \epsilon^2$ for
\begin{align}
T=O\left(\frac{\kappa^3}{p^2 \epsilon^2}\right) \mathcal{H}_0 \ell, \quad
K = O\left(\frac{p^2\sigma^2}{\kappa^2 n \epsilon^2} + \frac{\sigma^2}{\epsilon^2} + \frac{\kappa^2 \sigma^2}{n p \epsilon^2}\right),\nonumber
\end{align}
where  $\mathcal{H}_0=O\left(1+\frac{\delta_0}{K \kappa p}\right)$ and $\delta_0 = O\left(\frac{q}{\mu^2}\right)$.
\end{Theorem}

\begin{Remark}
    Focusing on the dependency of the convergence rate on accuracy $\epsilon$, the above theorem shows that in the regime of interest where $\epsilon$ gets small, the algorithm reaches an $\epsilon$-stationary point within $T=O(1/\epsilon^2)$ communication rounds, each consisting of $K = O(1/\epsilon^2)$ local updates. Therefore, the resulting SFO complexity is $T \cdot K = O(1/\epsilon^4)$. As we elaborated in Section II and Table \ref{comparison}, the proposed Dec-FedTrack algorithm simultaneously assembles all three components of local updates, heterogeneity and adversarial robustness.
\end{Remark}
\begin{Remark}
    It is also possible to derive the communication complexity for any given $K$. If we choose step-sizes $ \eta_c = \Theta(\frac{p}{\kappa^3 K \ell\sqrt{T}}) $, $ \eta_d = \Theta(\frac{p}{\kappa K \ell T}) $, and $\eta_s=\eta_r=\Theta(p)$, 
    after $T$ communication rounds each with $K$ local updates, there exists an iterate $0 \leq t \leq T$ such that $\mathbb{E}\|\nabla \Phi(\bar{\mathbf{x}}^{(t)})\|^2 \leq \epsilon^2$ for
    \begin{align}
    T = O\left(\frac{\kappa^6}{\epsilon^4 p^4} + \frac{p^4 \sigma^4}{n^2\kappa^4 K^2 \epsilon^4}+\frac{\kappa^4 \sigma^4}{n^2 K^2 p^2 \epsilon^4} \right)\nonumber,
\end{align}
which holds for any given $K$.
\end{Remark}

\subsection{Proof Sketch}

We first state the following standard results from optimization theory.

\begin{Proposition}\label{prop1}  Under Assumption \ref{assm: smooth}, $\Phi(\cdot) $ is $(\ell + \kappa \ell)$-smooth and $ \mathbf{y}^*(\cdot)=\arg \max _{\mathbf{y} \in \mathbb{R}^q} f(\cdot, \mathbf{y}) $ is $\kappa$-Lipschitz~\cite{lin2020gradient}.
\end{Proposition}
\begin{Proposition}\label{prop2}
Under Assumption \ref{assm: smooth}, for every $ \mathbf{x} \in \mathbb{R}^d$ and $ \mathbf{y}, \mathbf{y}' \in \mathbb{R}^q $, we have
\begin{align}
   \nabla_y f(\mathbf{x}, \mathbf{y})^{\top}\left(\mathbf{y}-\mathbf{y}^{\prime}\right)+\frac{1}{2 \ell}\left\|\nabla_y f(\mathbf{x}, \mathbf{y})\right\|^2+\frac{\mu}{2}\left\|\mathbf{y}-\mathbf{y}^{\prime}\right\|^2\leq f\left(\mathbf{x}, \mathbf{y}^{+}\right)-f\left(\mathbf{x}, \mathbf{y}'\right),  \nonumber
\end{align}
where $\mathbf{y}^{+}=\mathbf{y}-\frac{1}{\ell} \nabla_y f(\mathbf{x}, \mathbf{y})$~\cite{bubeck2015convex}.
\end{Proposition}

Next, we introduce some terminology that will be useful throughout the entire proof:
\begin{enumerate}
    \item The client (node) variance for variable $\mathbf{x}$ that measures the deviation of variable $\mathbf{x}$ at global steps from its averaged model:
$$
\Xi_t^x:=\frac{1}{n} \sum_i^n \mathbb{E}\|\mathbf{x}_i^{(t)}-\bar{\mathbf{x}}^{(t)}\|^2 .
$$
    \item Client-drift for variable $ \mathbf{x} $ that measures the deviation of the variable $\mathbf{x}$ at local steps from its averaged model:
$$
e_{k, t}^x:=\frac{1}{n} \sum_i^n \mathbb{E}\|\mathbf{x}_i^{(t)+k}-\bar{\mathbf{x}}^{(t)}\|^2 .
$$
The accumulation of local steps for  variable $\mathbf{x}$ is shown by
$$
\mathcal{E}_t^x:=\sum_{k=0}^{K-1} e_{k, t}^x=\sum_{k=0}^{K-1} \frac{1}{n} \sum_i^n \mathbb{E}\left\|\mathbf{x}_i^{(t)+k}-\bar{\mathbf{x}}^{(t)}\right\|^2 .
$$
    \item The quality of the correction for the variable $\mathbf{x}$ that measures the accuracy of the gradient correction in the local updates, which aims to bring local updates closer to global updates:
$$
\begin{aligned}
 \gamma_t^x=  \frac{1}{n \ell^2} \mathbb{E}\left\|\mathbf{C}^{(t)}+\nabla_x f\left(\bar{\mathbf{X}}^{(t)}, \bar{\mathbf{Y}}^{(t)}\right)-\nabla_x f\left(\bar{\mathbf{X}}^{(t)}, \bar{\mathbf{Y}}^{(t)}\right) \mathbf{J}\right\|_F^2.
\end{aligned}
$$
Similarly, we can define $\Xi_t^y, e_{k, t}^y, \mathcal{E}_t^y$, and $\gamma_t^y$ for variable $\mathbf{y}$.
    \item Consensus distance for variable $ \mathbf{y} $ that measures the deviation of the optimum $ \mathbf{y} $ when $ \mathbf{x} = \bar{\mathbf{x}}$ and the averaged $ \mathbf{y} $, that is, $\delta_t=\|\hat{\mathbf{y}}^{(t)}-\bar{\mathbf{y}}^{(t)}\|^2$
where $ \hat{\mathbf{y}}^{(t)}=\arg \max _{\mathbf{y} \in \mathbb{R}^q} f\left(\bar{\mathbf{x}}^{(t)}, \mathbf{y}\right).$
\end{enumerate}

Next, we provide recursion bounds for client variance, client drift, and quality of correction—for both variables $ \mathbf{x} $ and $ \mathbf{y} $—as well as consensus distance for variable $ \mathbf{y} $.
\begin{lemma}\label{lemma.1}
Under the assumption that $ \eta_c, \eta_d \lesssim \frac{1}{K\ell} $, we can bound the local drift for variables $\mathbf{x}$ and $\mathbf{y}$ as 
\begin{gather}
   \mathcal{E}_t^x  \lesssim K \Xi_t^x+ K^2 \eta_c^2 \ell^2 \mathcal{E}_t^y+ K^3 \eta_c^2 \ell^2 \gamma_t^x + K^3 \eta_c^2 \ell^2 \delta_t 
+ K^3 \eta_c^2 \mathbb{E}\left\|\nabla \Phi\left(\bar{\mathbf{x}}^{(t)}\right)\right\|^2 +K^2 \eta_c^2 \sigma^2,\nonumber \\
   \mathcal{E}_t^y \lesssim K \Xi_t^y+ K^2 \eta_d^2 \ell^2 \mathcal{E}_t^x+ K^3 \eta_d^2 \ell^2 \gamma_t^y + K^3 \eta_d^2 \ell^2 \delta_t + K^2 \eta_d^2 \sigma^2. \nonumber
\end{gather}
\end{lemma}
\begin{lemma}\label{lemma.2}
We have the following bounds on client variance for variable $ \mathbf{x} $ and $ \mathbf{y} $
\begin{align}
\Xi_{t+1}^x  &\lesssim\left(1-\frac{p}{2}\right)  \Xi_t^x+\frac{ K \eta_x^2 \ell^2}{p} \left(\mathcal{E}_t^x+\mathcal{E}_t^y\right)+\frac{ K^2 \eta_x^2 \ell^2}{p}  \gamma_t^x+ K \eta_x^2 \sigma^2,\nonumber\\
\Xi_{t+1}^y  &\lesssim\left(1-\frac{p}{2}\right)  \Xi_t^y+\frac{K \eta_y^2 \ell^2}{p} \left(\mathcal{E}_t^x+\mathcal{E}_t^y\right)+\frac{ K^2 \eta_y^2 \ell^2}{p}  \gamma_t^y+ K \eta_y^2 \sigma^2.\nonumber
\end{align}
\end{lemma}
\begin{lemma}\label{lemma.3}
Assuming that $ \eta_x, \eta_y\lesssim\frac{\sqrt{p}}{K \ell} $, we have the following bounds on the quality of correction for variables $ \mathbf{x} $ and $ \mathbf{y} $
\begin{align}
 \gamma_{t+1}^x &\lesssim\left(1-\frac{p}{2}\right) \gamma_t^x+\frac{1}{pK}\left(\mathcal{E}_t^x+\mathcal{E}_t^y\right)+\frac{ K^2 \ell^2}{p}\left(2 \eta_x^2+\eta_y^2\right) \delta_t
 +\frac{ K^2 \eta_x^2}{p}\mathbb{E}\left\|\nabla \Phi\left(\bar{\mathbf{x}}^{(t)}\right)\right\|^2+\frac{\sigma^2}{K \ell^2},\nonumber\\
 \gamma_{t+1}^y &\lesssim\left(1-\frac{p}{2}\right) \gamma_t^y+\frac{1}{pK}\left(\mathcal{E}_t^x+\mathcal{E}_t^y\right)+\frac{ K^2 \ell^2}{p}\left(2 \eta_x^2+\eta_y^2\right) \delta_t 
+\frac{ K^2 \eta_x^2}{p}\mathbb{E}\left\|\nabla \Phi\left(\bar{\mathbf{x}}^{(t)}\right)\right\|^2+\frac{\sigma^2}{K \ell^2}.\nonumber
\end{align}
\end{lemma}
\begin{lemma}\label{lemma.4}
Assuming that $ \eta_x \lesssim \frac{ \eta_y}{\kappa^2} $ and $ \eta_y \leq \frac{1}{K\ell} $, we have the following bound on $ \delta_t $ 
\begin{align}
 \delta_{t+1} &\lesssim \left(1-\frac{K \eta_y \ell}{\kappa}\right) \delta_t+\eta_y \ell\kappa \left(\mathcal{E}_t^x+\mathcal{E}_t^y\right)
+\frac{ \kappa^3 K \eta_x^2}{\eta_y \ell}\mathbb{E}\left\|\nabla \Phi\left(\bar{\mathbf{x}}^{(t)}\right)\right\|^2 +\frac{ \eta_y \sigma^2 \kappa}{n \ell}.   \nonumber
\end{align}
\end{lemma}
Now, we state the following descent lemma for $ \Phi(\mathbf{x}) $:
\begin{lemma}\label{lemma.5}
Assuming that $ \eta_x \lesssim \frac{1}{K\ell \kappa}$, we have the following bound on $ \mathbb{E} \Phi\left(\bar{\mathbf{x}}^{(t+1)}\right) $ as follows:
\begin{align}
    \mathbb{E} \Phi\left(\bar{\mathbf{x}}^{(t+1)}\right)&\lesssim \mathbb{E} \Phi\left(\bar{\mathbf{x}}^{(t)}\right)+ \eta_x \ell^2\left(\mathcal{E}_t^x+\mathcal{E}_t^y\right)+\ell^2 \eta_x K \delta_t-\eta_x K \mathbb{E}\left\|\nabla \Phi\left(\bar{\mathbf{x}}^{(t)}\right)\right\|^2+\frac{K \eta_x^2 \ell \sigma^2\kappa}{n}.\nonumber
\end{align}
\end{lemma}
Using Lemmas \ref{lemma.1}-\ref{lemma.5}, we have the following recursive bound on the Lyapunov function $ \mathcal{H}_t $.
\begin{lemma}\label{lemma.6}
Under the assumption that $ \eta_d = \Theta(\frac{p}{\kappa K\ell}) $, $ \eta_c = \Theta(\frac{\eta_d}{\kappa^2}) $, and $ \eta_s = \eta_r = \Theta(p) $, we can find constants $ A_x $, $ A_y $, $ B_x $, $ B_y $, and $ C $, such that 
\begin{align}
\mathcal{H}_{t+1}-\mathcal{H}_t & \lesssim -K \eta_x \mathbb{E}\left\|\nabla \Phi\left(\bar{\mathbf{x}}^{(t)}\right)\right\|^2+\frac{1}{p} K\ell \eta_d^3 \sigma^2+\frac{K \eta_x^2 \ell \kappa}{n} \sigma^2+\frac{\eta_y}{n p} \sigma^2,\nonumber
\end{align}
where
\begin{align}
   \mathcal{H}_t&=\mathbb{E} \Phi\left(\bar{\mathbf{x}}^{(t)}\right)-\mathbb{E} \Phi\left(\mathbf{x}^*\right)+A_x \eta_d K \ell^2 \Xi_t^x+A_y \eta_d K \ell^2 \Xi_t^y+B_x K^3 \ell^4 \eta_d^3 \gamma_t^x+B_y K^3 \ell^4 \eta_d^3 \gamma_t^y 
+C \frac{\ell}{ \kappa p} \delta_t.  \nonumber
\end{align}
\end{lemma}
Now, using the telescopic sum for $ \mathcal{H}_t $, we have
\begin{align}
 \frac{1}{T+1} \sum_{t=0}^T\left(\mathcal{H}_{t+1}-\mathcal{H}_t\right)&=\frac{1}{T+1}\left(\mathcal{H}_{T+1}-\mathcal{H}_0\right)\nonumber \\
& \lesssim- K \eta_x \frac{1}{T+1} \sum_{t=0}^T \mathbb{E}\left\|\nabla \Phi\left(\bar{\mathbf{x}}^{(t)}\right)\right\|^2+\frac{1}{p} K^2 \ell^2 \eta_d^3 \sigma^2+\frac{K \eta_x^2 \ell \kappa}{n} \sigma^2+\frac{\eta_y}{n  p} \sigma^2,\nonumber
\end{align}
which results in 
\begin{align}
 \frac{1}{T+1} \sum_{t=0}^T \mathbb{E}\left\|\nabla \Phi\left(\bar{\mathbf{x}}^{(t)}\right)\right\|^2 &\lesssim \frac{\mathcal{H}_0-\mathcal{H}_{T+1}}{(T+1)} \frac{1}{K \eta_x}+\frac{K \ell^2 \eta_d^3}{p\eta_x} \sigma^2+\frac{\eta_x \ell \kappa}{n } \sigma^2+\frac{\eta_y}{n K p \eta_x}. \label{thm}   
\end{align}
Now, we want to ensure $ \frac{1}{T+1} \sum_{t=0}^T \mathbb{E}\left\|\nabla \Phi\left(\bar{\mathbf{x}}^{(t)}\right)\right\|^2 \leq \epsilon^2 $ for any arbitrary $ \epsilon > 0 $, which is equivalent to aligning each term on the RHS of \eqref{thm} to the order of $ \epsilon^2 $. Given that $ \eta_x = \Theta\left(\frac{p^2}{\kappa^3 K \ell}\right) $, and $ \eta_y = \Theta\left(\frac{p^2}{\kappa K \ell}\right) $, we can conclude that 
$
T=O\left(\frac{\kappa^3}{p^2 \epsilon^2}\right) \mathcal{H}_0 \ell, \quad
K = O\left(\frac{p^2\sigma^2}{\kappa^2 n \epsilon^2} + \frac{\sigma^2}{\epsilon^2} + \frac{\kappa^2 \sigma^2}{n p \epsilon^2}\right).
$

\section{Empirical Results}
\subsection{Robust Logistic Regression}
\begin{figure*}[t]
    \centering 
    \scriptsize
    \begin{subfigure}[b]{0.24\textwidth} 
        \centering
    \includegraphics[width=\textwidth]{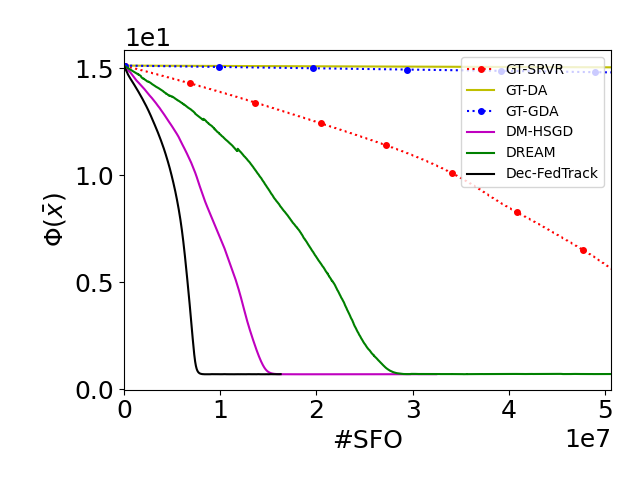}
        \caption{a9a}
    \end{subfigure}
    \hspace{0.03cm} 
    \begin{subfigure}[b]{0.24\textwidth}
        \centering
    \includegraphics[width=\textwidth]{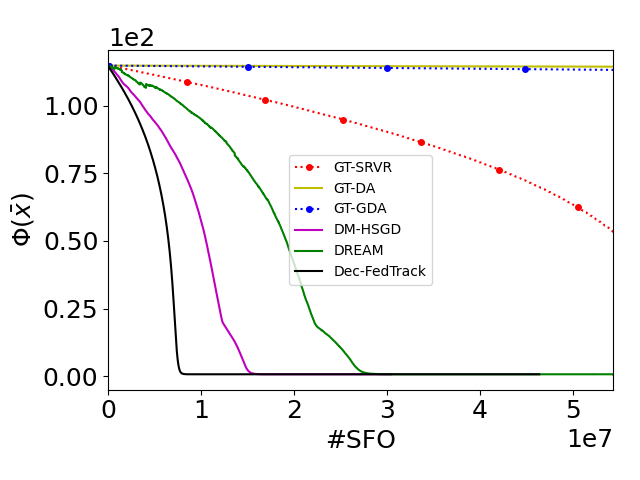}
        \caption{w8a}
    \end{subfigure}
    \hspace{0.03cm} 
    \begin{subfigure}[b]{0.24\textwidth}
        \centering
        \includegraphics[width=\textwidth]{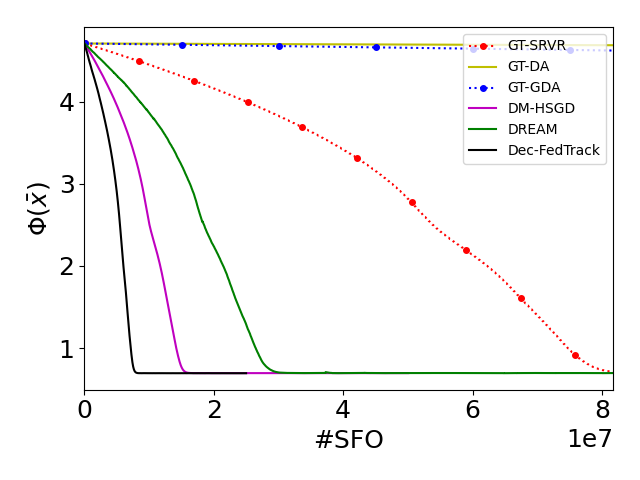}
        \caption{ijcnn1}
    \end{subfigure}
    \hspace{0.03cm}
    \begin{subfigure}[b]{0.24\textwidth}
        \centering
        \includegraphics[width=\textwidth]{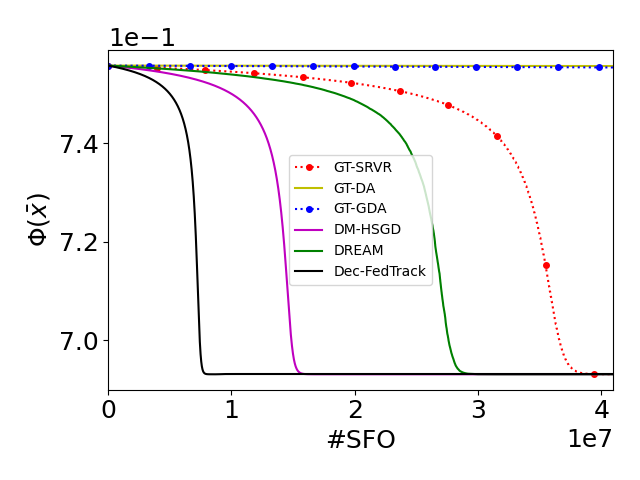}
        \caption{phishing}
    \end{subfigure}
    
    \vspace{0.1cm} 
    
    \begin{subfigure}[b]{0.24\textwidth}
        \centering
        \includegraphics[width=\textwidth]{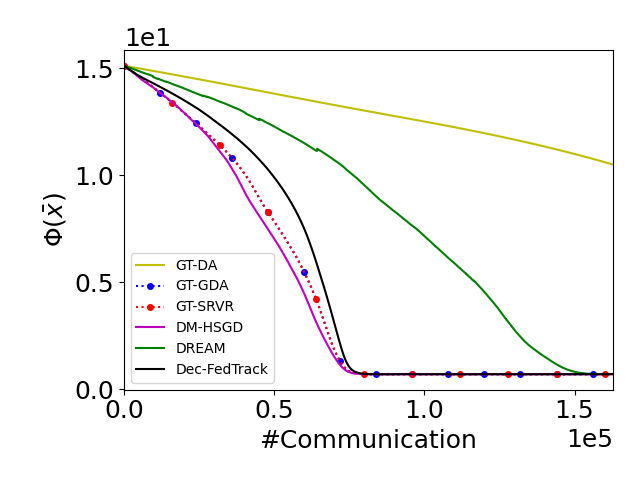}
        \caption{a9a}
    \end{subfigure}
    \hspace{0.03cm} 
    \begin{subfigure}[b]{0.24\textwidth}
        \centering
        \includegraphics[width=\textwidth]{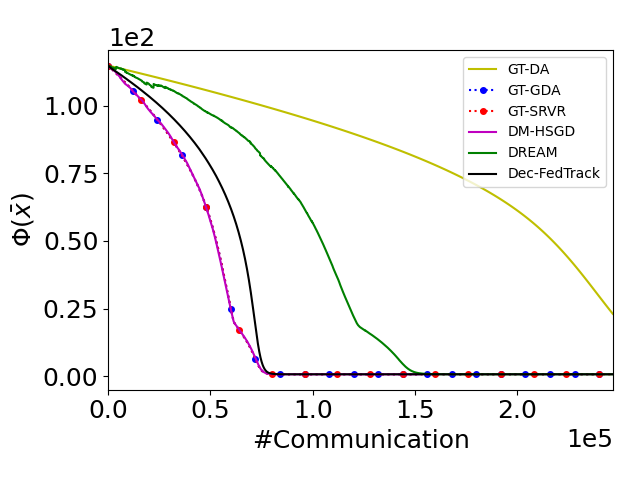} 
        \caption{w8a}
    \end{subfigure}
    \hspace{0.03cm} 
    \begin{subfigure}[b]{0.24\textwidth}
        \centering
        \includegraphics[width=\textwidth]{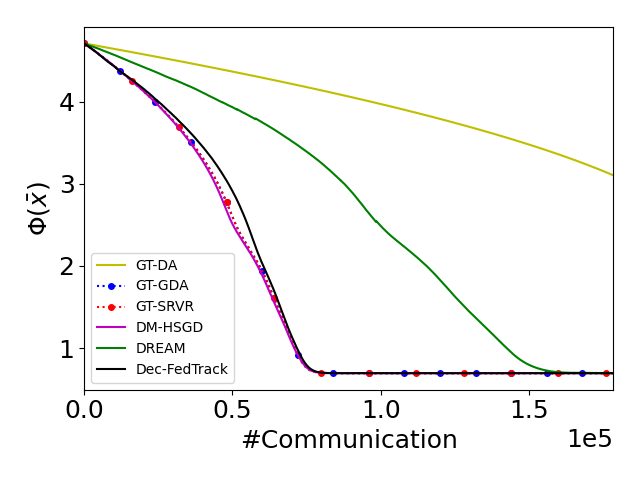} 
        \caption{ijcnn1}
    \end{subfigure}
    \hspace{0.03cm}
    \begin{subfigure}[b]{0.24\textwidth}
        \centering
        \includegraphics[width=\textwidth]{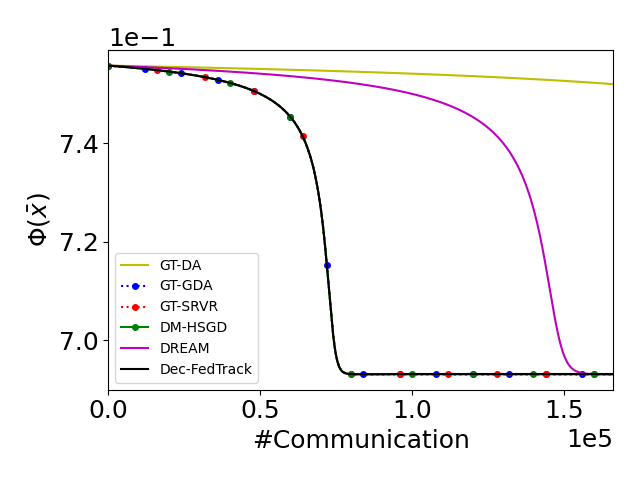} 
        \caption{phishing}
    \end{subfigure}

    \caption{Convergence of $\Phi(\bar{\mathbf{x}})$ against the number of SFO calls (above) and the number of communication rounds (bottom).}
    \label{fig1}
\end{figure*}
    
We consider the problem of training a robust logistic regression classifier with a non-convex regularizer similar to \cite{xian2021faster,chen2022simple,luo2020stochastic}. In this problem, we aim to train a binary classifier $ \mathbf{x} \in \mathbb{R}^d $ on the dataset $ \left\{\left(a_{ij}, b_{ij}\right)\right\} $, where $ 
a_{ij} \in \mathbb{R}^d $ denotes the feature vector and $ b_{ij} \in \{-1, +1\} $ represents the label for the $j$th sample in the dataset associated with client $i$. Each client is allocated $ m $ samples, resulting in a total of $ N = mn $ samples.  The loss function at  client $i$ is given by
\begin{align}
    f_i(\mathbf{x},\mathbf{y}) \triangleq \frac{1}{m} \sum_{j=1}^m\left(\mathbf{y}_{i j} l_{i j}(\mathbf{x})-V(\mathbf{y})+g(\mathbf{x})\right),\nonumber
\end{align}
where $l_{i j}(\mathbf{x})=\log \left(1+\exp \left(b_{i j} a_{i j}^{\top} \mathbf{x}\right)\right)$, $V(\mathbf{y})=\frac{1}{2 N^2}\|N \mathbf{y}-\mathbf{1}\|^2$,  $g(\mathbf{x})=\theta \sum_{k=1}^d \frac{\nu x_k^2}{1+\nu x_k^2}$, $\theta=10^{-5}$, and $\nu=10$. The parameter $\mathbf{y}$ is restricted to the simplex $\Delta_N=\{\mathbf{y} \in \mathbb{R}^N: y_k \in[0,1], \sum_{k=1}^N y_k=1 \}$. Here, we set the mixing matrix $\mathbf{W}$ as the $\pi$-lazy random walk matrix~\cite{xian2021faster} on a ring graph with $n = 10$. 

As previously highlighted, the main distinction of Dec-FedTrack compared to other decentralized minimax methods lies in its use of multiple local updates, which aligns well with FL applications. Notably, multiple local steps are essential in FL to ensure privacy. 

However, in this section, we set the number of local updates for the Dec-FedTrack algorithm to $1$ ($K = 1$) and compare our proposed algorithm against DREAM~\cite{chen2022simple}, DM-HSGD~\cite{xian2021faster}, GT-DA~\cite{tsaknakis2020decentralized}, GT-GDA, and GT-SRVR~\cite{zhang2021taming}. These comparisons are conducted on the datasets ``a9a'', ``ijcnn1'', ``phishing'', and ``w8a''~\cite{chang2011libsvm}, evaluating performance in terms of the number of SFO calls and communication rounds against $\Phi(\bar{\mathbf{x}}) = \max_{\mathbf{y} \in \Delta_N} f(\bar{\mathbf{x}}, \mathbf{y})$, as well as test accuracy. 

\begin{figure*}[t]
    \centering 
    \scriptsize
    \begin{subfigure}[b]{0.3\textwidth} 
        \centering
    \includegraphics[width=\textwidth]{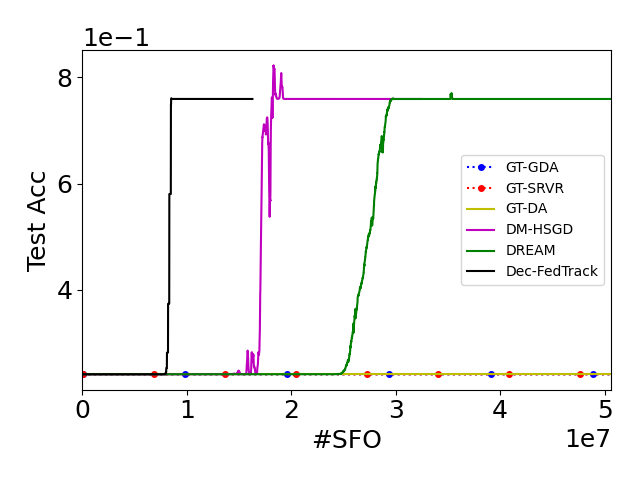}
        \caption{a9a}
    \end{subfigure}
    \hspace{0.05cm} 
    \begin{subfigure}[b]{0.3\textwidth}
        \centering
    \includegraphics[width=\textwidth]{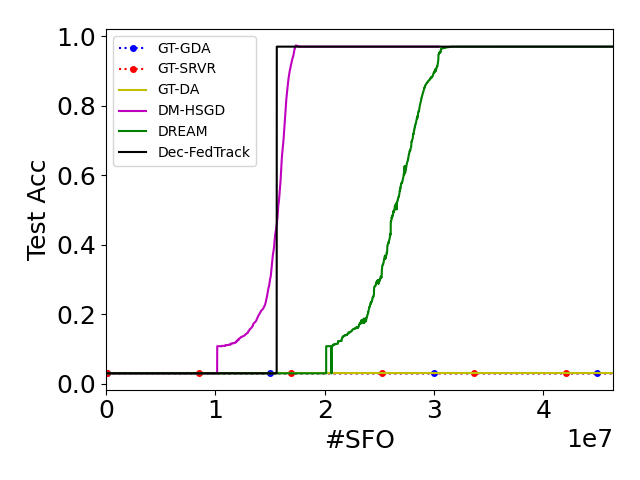}
        \caption{w8a}
    \end{subfigure}
    \hspace{0.05cm} 
    \begin{subfigure}[b]{0.3\textwidth}
        \centering
        \includegraphics[width=\textwidth]{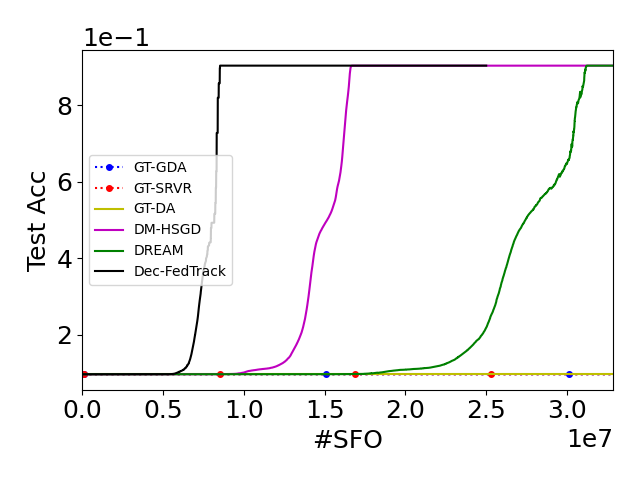}
        \caption{ijcnn1}
    \end{subfigure}
    
    \vspace{0.1cm} 
    
    \begin{subfigure}[b]{0.3\textwidth}
        \centering
        \includegraphics[width=\textwidth]{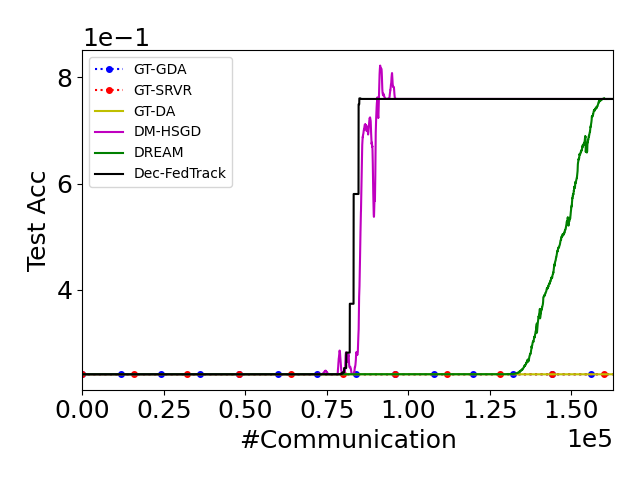}
        \caption{a9a}
    \end{subfigure}
    \hspace{0.05cm} 
    \begin{subfigure}[b]{0.3\textwidth}
        \centering
        \includegraphics[width=\textwidth]{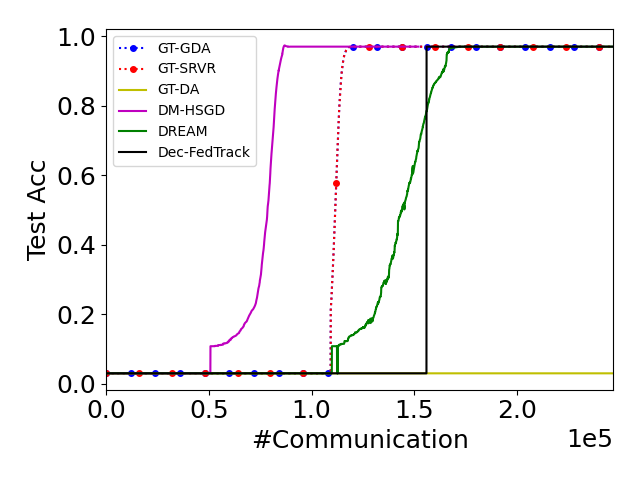} 
        \caption{w8a}
    \end{subfigure}
    \hspace{0.05cm} 
    \begin{subfigure}[b]{0.3\textwidth}
        \centering
        \includegraphics[width=\textwidth]{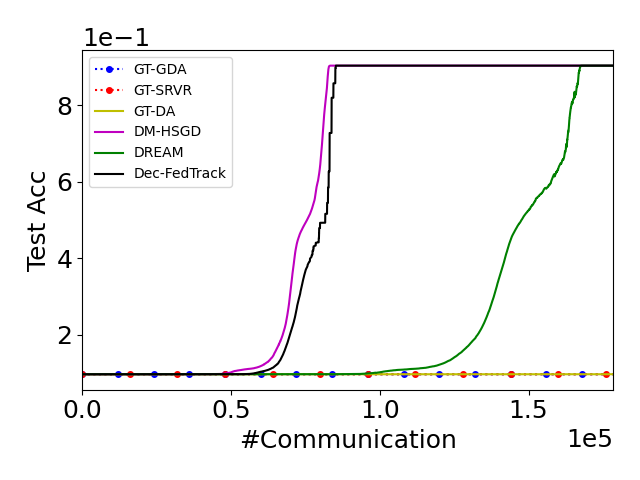} 
        \caption{ijcnn1}
    \end{subfigure}

    \caption{Test accuracy against the number of SFO calls (above) and the number of communication rounds (bottom). }
    \label{fig2}
\end{figure*}

We fix the batch size to $64$ across all algorithms and tune the learning rates with $ \eta_x \in \{0.1, 0.01, 0.001, 0.0001\} $ and $ \eta_y \in \{1, 0.1, 0.01, 0.001\} $. Fig.~\ref{fig1} presents the comparison of the number of SFO calls and number of communication rounds against $\Phi(\bar{\mathbf{x}})$ on datasets ``a9a'', ``ijcnn1'', ``phishing'', and ``w8a''. As shown, Dec-FedTrack demonstrates a faster decay rate on $\Phi(\bar{\mathbf{x}})$ against the number of SFO calls and faster or very close decay rate on $\Phi(\bar{\mathbf{x}})$ against the number of communications. Furthermore, Fig. \ref{fig2} compares the comparison of the number of SFO calls and number of communication rounds against the test accuracy on datasets ``a9a'', ``ijcnn1'', and ``w8a''. Note that the ``phishing'' dataset does not include a test dataset.
 
\subsection{Robust Neural Network Training}
\begin{table*}[t]
\centering
\caption{Test accuracy for K-GT and Dec-FedTrack algorithms under different attack methods and adversary budgets.}
\label{tab:robust_nn_results}
\resizebox{\textwidth}{!}{%
\begin{tabular}{@{}clccccccccccc@{}}
\toprule
\multicolumn{1}{c}{  Dataset \& Model } &
\multicolumn{1}{l}{Method} & \multirow{1}{*}{Clean Acc.} & \multicolumn{3}{c}{$\text{FGSM}\;L_{\infty}$~\cite{goodfellow2014explaining}} & \multicolumn{3}{c}{$\text{PGD}\; L_{\infty}$~\cite{kurakin2016adversarial}} &
\multicolumn{3}{c}{UAP~\cite{mummadi2019defending}} &\\
\midrule
 &  &  & $\delta=0.05$ & $\delta=0.1$ & $\delta=0.15$ & $\delta=0.05$ & $\delta=0.1$ & $\delta=0.15$ & $\delta=0.20$ & $\delta=0.25$ & $\delta=0.30$ \\
 \cmidrule(l){4-12}
 \multirow{1}{*}{ MNIST  } & K-GT & $\bm{99.20}$ & $93.73$ & $73.10$ & $39.65$ & $94.90$ & $74.67$ & $30.86$ & $93.64$ & $75.15$ & $36.26$ \\
 & Dec-FedTrack & $99.14$ &  $\bm{94.83}$ & $\bm{78.02}$ & $\bm{49.06}$ & $\bm{96.20}$ & $\bm{81.72}$ & $\bm{46.49}$& $\bm{96.14}$ & $\bm{85.73}$ & $\bm{43.87}$    \\
 \midrule
 &  &  & $\delta=0.003$ & $\delta=0.005$ & $\delta=0.01$ & $\delta=0.003$ & $\delta=0.005$ & $\delta=0.01$ & $\delta=0.03$ & $\delta=0.05$ & $\delta=0.07$ \\
 \cmidrule(l){4-12} 
 \multirow{1}{*}{ CIFAR-10 } & K-GT & \bm{$77.3$} & $67.7$ & $44.8$ & $23.6$ & $67.6$ & $44.6$ & $26.4$ & $58.9$ & $53.3$ & $51.5$\\
  & Dec-FedTrack & $77.1$ & \bm{$69.7$} & \bm{$51.5$} & \bm{$32.5$} & \bm{$69.5$} & \bm{$51.7$} & \bm{$35.9$} & \bm{$74.9$} & \bm{$66.1$} & \bm{$56.3$} \\
\bottomrule
\end{tabular}%
}
\end{table*}

In this section, we consider the problem of robust neural network (NN) training, in the presence of adversarial perturbations. We consider a similar problem as considered in~\cite{deng2020distributionally},
$$
\min_{\mathbf{x}} \max _{\|\mathbf{y}\|_{\infty} \leq \delta} \frac{1}{n} \sum_{i=1}^n f_i(\mathbf{x},\mathbf{y})
$$
where $f_i(\mathbf{x},\mathbf{y}) \coloneqq 1/m \sum_{j=1}^m \ell\left(h_{\mathbf{x}}\left(a_{ij}+\mathbf{y}\right), b_{ij}\right)$. Here, $\mathbf{x}$ denotes the parameters of the $\mathrm{NN}, \mathbf{y}$ denotes the perturbation, and $\left(a_{ij}, b_{ij}\right)$ denotes the $j$-th data sample of client $i$.

We consider the accuracy of our formulation against three popular attacks: The Fast Gradient Sign Method (FGSM)~\cite{goodfellow2014explaining}, Projected Gradient descent (PGD)~\cite{kurakin2016adversarial}, and Universal Adversarial Perturbation (UAP)~\cite{shafahi2020universal}. We have provided a description of each attack in Appendix~\ref{appendix2}.

We evaluate the robustness of Dec-FedTrack against adversarial attacks by comparing it with K-GT, a benchmark minimization algorithm. The evaluation was conducted on the MNIST and CIFAR-10 datasets, utilizing 2-layer and 3-layer convolutional neural networks for training MNIST and CIFAR-10, respectively. For CIFAR-10 experiments, we only use two classes to demonstrate the efficacy of our method.

During training, we set $n = 5$, $K = 5$, and experiment with various constant learning rates chosen from $\{1, 0.5, 0.1, 0.05, 0.01\}$, using a batch size of $128$. The results for K-GT and our proposed algorithm under different attack methods and varying values of $\delta$ are summarized in Table~\ref{tab:robust_nn_results}. As shown in the table, the proposed algorithm demonstrates superior performance compared to its non-robust counterpart.

\section{Conclusion}
This paper presents Dec-FedTrack, a decentralized minimax optimization algorithm specifically tailored for addressing the challenges prevalent in distributed learning systems, particularly within federated learning setups. Dec-FedTrack, by integrating local updates and gradient tracking mechanisms, aims to enhance robustness against universal adversarial perturbations while efficiently mitigating data heterogeneity. The theoretical analysis establishes convergence guarantees under certain assumptions, affirming Dec-FedTrack's reliability and efficacy. Our empirical evaluations demonstrate that for an equal adversary budget, Dec-FedTrack is more robust to adversarial perturbations compared to non-robust baselines such as K-GT.

\bibliographystyle{ieeetr} 
\bibliography{main.bib}

\appendix


\section{Appendix}
\newtheorem{assumptionalt}{Lemma}[lemma]
\newenvironment{lemmap}[1]{
  \renewcommand\theassumptionalt{#1}
  \assumptionalt
}{\endassumptionalt}

\subsection{Proof of Intermediate Lemmas}
\begin{lemmap}{A.1}\label{lem.1} For a set of arbitrary vectors $ a_1, \ldots, a_n $ such that $ a_i \in \mathbb{R}^d $, we have
\begin{align}
    \left\|\frac{1}{n}\sum_{i=1}^n a_i \right\|^2 \leq \frac{1}{n}\sum_{i=1}^n\|a_i\|^2.\nonumber
\end{align}
\end{lemmap}
\begin{lemmap}{A.2}\label{lem.2} (Young's Inequality) For any vectors $ a, b \in \mathbb{R}^d$ and $\alpha > 0$ we have
\begin{gather}
    2\langle a, b\rangle \leq \alpha\|a\|^2+\frac{1}{\alpha}\|b\|^2,\nonumber\\
    \|a + b\|^2 \leq (1 + \alpha)\|a\|^2 + (1 + \frac{1}{\alpha})\|b\|^2.\nonumber
\end{gather}
\end{lemmap}
\begin{lemmap}{A.3}\label{lem.3}
If we initialize $ \mathbf{C}^{(0)} $ and $ \mathbf{D}^{(0)} $ as below
\begin{align}
    \mathbf{c}_i^{(0)}=-\nabla_x F_i\left(\mathbf{x}^{(0)}, \mathbf{y}^{(0)} ; \xi_i\right)+\frac{1}{n} \sum_j \nabla_x F_j\left(\mathbf{x}^{(0)}, \mathbf{y}^{(0)} ; \xi_j\right), \nonumber\\
    \mathbf{d}_i^{(0)}=-\nabla_y F_i\left(\mathbf{x}^{(0)}, \mathbf{y}^{(0)} ; \xi_i\right)+\frac{1}{n} \sum_j \nabla_y F_j\left(\mathbf{x}^{(0)}, \mathbf{y}^{(0)} ; \xi_j\right),  \label{eq.0} 
\end{align}  
then the averaged correction for variables $ \mathbf{x} $ and $ \mathbf{y} $ in any communication round equals to zero.
\end{lemmap}
\textit{Proof.} According to Algorithm \ref{al.1} we have
\begin{align}
    \mathbf{C}^{(t+1)} \mathbf{J}=\mathbf{C}^{(t)} \mathbf{J}+\frac{1}{K \eta_c}\left(\mathbf{X}^{(t)}-\mathbf{X}^{(t)+K}\right)(\mathbf{W}-\mathbf{I}) \mathbf{J}=\mathbf{C}^{(t)} \mathbf{J}.\nonumber
\end{align}
Using the initialization assumption in \eqref{eq.0}, we have $ \mathbf{C}^{(t)} \mathbf{J}=\mathbf{C}^{(0)} \mathbf{J}=\mathbf{0} $. Similarly, we have $ \mathbf{D}^{(t)} \mathbf{J}=\mathbf{D}^{(0)} \mathbf{J}=\mathbf{0} $.
\hfill \ensuremath{\Box} 


\begin{lemmap}{A.4} \label{lem.4} Using Assumption \ref{assm: smooth} and Young's Inequality we have 
\begin{gather}
\mathbb{E}\left\|\nabla_x f\left(\bar{\mathbf{x}}^{(t)}, \bar{\mathbf{y}}^{(t)}\right)\right\|^2 \leq 2 \ell^2 \delta_t+2\mathbb{E}\left\|\nabla \Phi\left(\bar{\mathbf{x}}^{(t)}\right)\right\|^2,\nonumber\\
\mathbb{E}\left\|\nabla_y f\left(\bar{\mathbf{x}}^{(t)}, \bar{\mathbf{y}}^{(t)}\right)\right\|^2 \leq \ell^2 \delta_t.\nonumber
\end{gather}
\end{lemmap}
\textit{Proof.} We can write
\begin{align}
\mathbb{E}\left\|\nabla_x f\left(\bar{\mathbf{x}}^{(t)}, \bar{\mathbf{y}}^{(t)}\right)\right\|^2 & =\mathbb{E}\left\|\nabla_x f\left(\bar{\mathbf{x}}^{(t)}, \bar{\mathbf{y}}^{(t)}\right)-\nabla_x f\left(\bar{\mathbf{x}}^{(t)}, \hat{\mathbf{y}}^{(t)}\right) +\nabla_x f\left(\bar{\mathbf{x}}^{(t)}, \hat{\mathbf{y}}^{(t)}\right)\right\|^2 \nonumber\\&
 \leq 2 \ell^2\mathbb{E}\left\|\bar{\mathbf{y}}^{(t)}-\hat{\mathbf{y}}^{(t)}\right\|^2+2\mathbb{E}\left\|\nabla \Phi\left(\bar{\mathbf{x}}^{(t)}\right)\right\|^2=2 \ell^2 \delta_t+2\mathbb{E}\left\|\nabla \Phi\left(\bar{\mathbf{x}}^{(t)}\right)\right\|^2.\nonumber
\end{align}
Moreover,
\begin{align}\label{eq.1}
    &\mathbb{E}\left\|\nabla_y f\left(\bar{\mathbf{x}}^{(t)}, \bar{\mathbf{y}}^{(t)}\right)\right\|^2  
=\mathbb{E}\left\|\nabla_y f\left(\bar{\mathbf{x}}^{(t)}, \bar{\mathbf{y}}^{(t)}\right)-\nabla_y f\left(\bar{\mathbf{x}}^{(t)}, \hat{\mathbf{y}}^{(t)}\right)\right\|^2 \leq \ell^2\delta_t.  
\end{align}
The equality in \eqref{eq.1} holds due to the fact that $\nabla_y f\left(\bar{\mathbf{x}}^{(t)}, \hat{\mathbf{y}}^{(t)}\right)\ = 0$.
\hfill \ensuremath{\Box} 
\begin{lemmap}{A.5}\label{lem.5} 
Under the assumption that $ \eta_c, \eta_d \leq \frac{1}{8K\ell} $, we can bound the local drift for variables $\mathbf{x}$ and $\mathbf{y}$ as follows
\begin{gather}
   \mathcal{E}_t^x  \leq 3 K \Xi_t^x+12 K^2 \eta_c^2 \ell^2 \mathcal{E}_t^y+12 K^3 \eta_c^2 \ell^2 \gamma_t^x+12 K^3 \eta_c^2 \ell^2 \delta_t + 12 K^3 \eta_c^2 \mathbb{E}\left\|\nabla \Phi\left(\bar{\mathbf{x}}^{(t)}\right)\right\|^2 +3 K^2 \eta_c^2 \sigma^2, \nonumber\\
   \mathcal{E}_t^y  \leq 3 K \Xi_t^y+12 K^2 \eta_d^2 \ell^2 \mathcal{E}_t^x+12 K^3 \eta_d^2 \ell^2 \gamma_t^y +6 K^3 \eta_d^2 \ell^2 \delta_t +3 K^2 \eta_d^2 \sigma^2. \nonumber
\end{gather}
\end{lemmap}
\textit{Proof.}
For $ K =1 $ the inequalities obviously hold since $ \mathcal{E}^x_t = \Xi^x_t = \frac{1}{n} \mathbb{E}\left\|\mathbf{X}^{(t)}-\bar{\mathbf{X}}^{(t)}\right\|_F^2  $ and $ \mathcal{E}^y_t = \Xi^y_t = \frac{1}{n} \mathbb{E}\left\|\mathbf{Y}^{(t)}-\bar{\mathbf{Y}}^{(t)}\right\|_F^2  $ and other terms on the RHSs are positive. For $ K \geq 2 $ we have
\begin{equation}
\begin{aligned}
 n e_{k, t}^x & :=\mathbb{E}\left\|\mathbf{X}^{(t)+k}-\bar{\mathbf{X}}^{(t)}\right\|_F^2 \\
&= \mathbb{E}\left\|\mathbf{X}^{(t)+k-1}-\eta_c\left(\nabla_x F\left(\mathbf{X}^{(t)+k-1}, \mathbf{Y}^{(t)+k-1} ; \xi^{(t)+k-1}\right)+\mathbf{C}^{(t)}\right)-\bar{\mathbf{X}}^{(t)}\right\|_F^2 \\
& \leq\left(1+\frac{1}{K-1}\right) \mathbb{E}\left\|\mathbf{X}^{(t)+k-1}-\bar{\mathbf{X}}^{(t)}\right\|^2+n \eta_c^2 \sigma^2 +K \eta_c^2 \mathbb{E} \left\| \nabla_x f\left(\mathbf{X}^{(t)+k-1}, \mathbf{Y}^{(t)+k-1}\right)\right.\\
&\quad \left. -\nabla_x f\left(\bar{\mathbf{X}}^{(t)}, \bar{\mathbf{Y}}^{(t)}\right) +\mathbf{C}^{(t)}+\nabla_x f\left(\bar{\mathbf{X}}^{(t)}, \bar{\mathbf{Y}}^{(t)}\right)(\mathbf{I}-\mathbf{J})+\nabla_x f\left(\bar{\mathbf{X}}^{(t)}, \bar{\mathbf{Y}}^{(t)}\right) \mathbf{J} \right\|_F^2 \\
& \leq \underbrace{\left(1+\frac{1}{K-1}+4 K \eta_c^2 \ell ^2\right)}_{:=\mathcal{C}} \mathbb{E}\left\|\mathbf{X}^{(t)+k-1}-\bar{\mathbf{X}}^{(t)}\right\|_F^2+4 K \eta_c^2 \ell^2 \mathbb{E}\left\|\mathbf{Y}^{(t)+k-1}-\bar{\mathbf{Y}}^{(t)}\right\|_F^2 \\
& \quad +4 K \eta_c^2 \ell^2 n \gamma_t^x+2 K \eta_c^2 n \mathbb{E}\left\|\nabla_x f\left(\bar{\mathbf{x}}^{(t)}, \bar{\mathbf{y}}^{(t)}\right)\right\|^2+n \eta_c^2 \sigma^2 \\
& \leq \mathcal{C}^k \mathbb{E}\left\|\mathbf{X}^{(t)}-\bar{\mathbf{X}}^{(t)}\right\|_F^2 +\sum_{r=0}^{k-1} \mathcal{C}^r\left(4 K \eta_c^2 \ell^2 \mathbb{E}\left\|\mathbf{Y}^{(t)+k-r-1}-\bar{\mathbf{Y}}^{(t)}\right\|_F^2\right.  \left.+4 K \eta_c^2 \ell^2 n \gamma_t^x\right.\\
& \quad \left.+2 K \eta_c^2 n \mathbb{E}\left\|\nabla_x f\left(\bar{\mathbf{x}}^{(t)}, \bar{\mathbf{y}}^{(t)}\right)\right\|^2+n \eta_c^2 \sigma^2\right) \nonumber
\end{aligned}
\end{equation}
If the condition $\eta_c \leq \frac{1}{8 K \ell}$ holds, then it follows that $4 K\left(\eta_c \ell\right)^2 \leq \frac{1}{16 K}<\frac{1}{16(K-1)}$. Given $\mathcal{C}>1$, it can be established that $\mathcal{C}^k \leq \mathcal{C}^K \leq\left(1+\frac{1}{K-1}+\frac{1}{16(K-1)}\right)^K \leq e^{1+\frac{1}{16}} \leq 3$.
Now, we can obtain a bound on client drift for variable $\mathbf{x}$

\begin{align}\label{eq.2}
\mathcal{E}_t^x= & \sum_{k=0}^{K-1} e_{k, t}^x \leq 3 K \Xi_t^x+12 K^2 \eta_c^2 \ell^2 \mathcal{E}_t^y+12 K^3 \eta_c^2 \ell^2 \gamma_t^x +6 K^3 \eta_c^2 \mathbb{E}\left\|\nabla_x f\left(\bar{\mathbf{x}}^{(t)}, \bar{\mathbf{y}}^{(t)}\right)\right\|^2+3 K^2 \eta_c^2 \sigma^2.
\end{align}
Similarly, a bound on client drift for variable $\mathbf{y}$ can be formulated by
\begin{align}\label{eq.3}
\mathcal{E}_t^y= & \sum_{k=0}^{K-1} e_{k, t}^y \leq 3 K \Xi_t^y+12 K^2 \eta_d^2 \ell^2 \mathcal{E}_t^x+12 K^3 \eta_d^2 \ell^2 \gamma_t^y +6 K^3 \eta_d^2 \mathbb{E}\left\|\nabla_y f\left(\bar{\mathbf{x}}^{(t)}, \bar{\mathbf{y}}^{(t)}\right)\right\|^2+3 K^2 \eta_d^2 \sigma^2.
\end{align}
Using Lemma \ref{lem.4} in \eqref{eq.2} and \eqref{eq.3} will complete the proof.
\hfill \ensuremath{\Box} 

\begin{lemmap}{A.6}\label{lem.6}
We have the following bounds on client variance for variable $ \mathbf{x} $ and $ \mathbf{y} $
\begin{align}
\Xi_{t+1}^x  &\leq\left(1-\frac{p}{2}\right)  \Xi_t^x+\frac{6 K \eta_x^2 \ell^2}{p} \left(\mathcal{E}_t^x+\mathcal{E}_t^y\right)+\frac{6 K^2 \eta_x^2 \ell^2}{p}  \gamma_t^x+ K \eta_x^2 \sigma^2,\nonumber\\
\Xi_{t+1}^y  &\leq\left(1-\frac{p}{2}\right)  \Xi_t^y+\frac{6 K \eta_y^2 \ell^2}{p} \left(\mathcal{E}_t^x+\mathcal{E}_t^y\right)+\frac{6 K^2 \eta_y^2 \ell^2}{p}  \gamma_t^y+ K \eta_y^2 \sigma^2.\nonumber
\end{align}
\end{lemmap}
\textit{Proof.} Using the update rule $ \mathbf{X}^{(t+1)} =  \mathbf{X}^{(t)}-\eta_x \sum_{k=0}^{K-1}\left(\nabla_x F\left(\mathbf{X}^{(t)+k}, \mathbf{Y}^{(t)+k} ; \xi^{(t)+k}\right)+\mathbf{C}^{(t)}\right)$ derived from Algorithm \ref{al.1}, we can bound the client variance for variable $ \mathbf{x} $
\begin{equation}
\begin{aligned}
 n \Xi_{t+1}^x &=\mathbb{E}\left\|\mathbf{X}^{(t+1)}-\bar{\mathbf{X}}^{(t+1)}\right\|^2 \\
& =\mathbb{E}\left\|\left(\mathbf{X}^{(t)}-\eta_x \sum_{k=0}^{K-1}\left(\nabla_x F\left(\mathbf{X}^{(t)+k}, \mathbf{Y}^{(t)+k} ; \xi^{(t)+k}\right)+\mathbf{C}^{(t)}\right)\right)(\mathbf{W}-\mathbf{J})\right\|_F^2 \\
& \stackrel{(a)}{\leq}(1-p) \mathbb{E}\left\|\left(\mathbf{X}^{(t)}-\eta_x \sum_{k=0}^{K-1}\left(\nabla_x f\left(\mathbf{X}^{(t)+k}, \mathbf{Y}^{(t)+k}\right)+\mathbf{C}^{(t)}\right)\right)(\mathbf{I}-\mathbf{J})\right\|_F^2+n K \eta_x^2 \sigma^2 \\
& \leq n K \eta_x^2 \sigma^2+(1+\alpha)(1-p) \mathbb{E}\left\|\mathbf{X}^{(t)}(\mathbf{I}-\mathbf{J})\right\|_F^2 \\
& \quad +\left(1+\frac{1}{\alpha}\right) \eta_x^2 \mathbb{E} \left\| \sum_{k=0}^{K-1} \nabla_x f\left(\mathbf{X}^{(t)+k}, \mathbf{Y}^{(t)+k}\right)(\mathbf{I}-\mathbf{J})-K \nabla_x f\left(\bar{\mathbf{X}}^{(t)}, \bar{\mathbf{Y}}^{(t)}\right)(\mathbf{I}-\mathbf{J})\right. \\
& \quad\left. +K \nabla_x f\left(\bar{\mathbf{X}}^{(t)}, \bar{\mathbf{Y}}^{(t)}\right)(\mathbf{I}-\mathbf{J})+K \mathbf{C}^{(t)} \right\|_F^2 \\
& \stackrel{\alpha = \frac{p}{2}, p \leq 1}{\leq} n K \eta_x^2 \sigma^2+\left(1-\frac{p}{2}\right) \mathbb{E}\left\|\mathbf{X}^{(t)}-\bar{\mathbf{X}}^{(t)}\right\|_F^2 \\
& \quad +\frac{6}{p}\left(K \eta_x^2 \ell^2\|\mathbf{I}-\mathbf{J}\|^2\left(\sum_{k=0}^{K-1}\left\|\mathbf{X}^{(t)+k}-\bar{\mathbf{X}}^{(t)}\right\|_F^2+\sum_{k=0}^{K-1} \mathbb{E}\left\|\mathbf{Y}^{(t)+k}-\bar{\mathbf{Y}}^{(t)}\right\|_F^2\right)\right. \\
& \quad \left.+K^2 \eta_x^2 \mathbb{E}\left\|\nabla_x f\left(\bar{\mathbf{X}}^{(t)}, \bar{\mathbf{Y}}^{(t)}\right)(\mathbf{I}-\mathbf{J})+\mathbf{C}^{(t)}\right\|_F^2\right) \\
& \leq\left(1-\frac{p}{2}\right) n \Xi_t^x+\frac{6 K \eta_x^2 \ell^2}{p} n\left(\mathcal{E}_t^x+\mathcal{E}_t^y\right)+\frac{6 K^2 \eta_x^2 \ell^2}{p} n \gamma_t^x+n K \eta_x^2 \sigma^2.\nonumber
\end{aligned}
\end{equation}
where we used Assumption \ref{assm: W} in $ (a) $. Similarly, we can derive an upper bound on client variance for variable $ \mathbf{y} $, thereby concluding the proof.
\hfill \ensuremath{\Box} 
\begin{lemmap}{A.7}\label{lem.7}
The sum of averaged progress between communications for variables $\mathbf{x}$ and $ \mathbf{y} $ can be bounded by
\begin{align}
\Delta_{t+1}^x+\Delta_{t+1}^y  &\leq 2 K \ell^2\left(\eta_x^2+\eta_y^2\right)\left(\mathcal{E}_t^x+\mathcal{E}_t^y\right)+2 K^2 \ell^2\left(2 \eta_x^2+\eta_y^2\right) \delta_t+4 K^2 \eta_x^2 \mathbb{E}\left\|\nabla \Phi\left(\bar{x}^{(t)}\right)\right\|^2 \nonumber\\&\quad+ \frac{K \sigma^2}{n}\left(\eta_x^2+\eta_y^2\right).\nonumber
\end{align}
\end{lemmap}
\textit{Proof.}
First, we derive an upper bound on the averaged progress for variable $ \mathbf{x} $ as follows
\begin{align}
 \Delta_{t+1}^x &:=\mathbb{E}\left\|\bar{\mathbf{x}}^{(t+1)}-\bar{\mathbf{x}}^{(t)}\right\|^2 \nonumber\\
& =\eta_x^2 \mathbb{E}\left\|\frac{1}{n} \sum_{i, k} \nabla_x F_i\left(\mathbf{x}_i^{(t)+k}, \mathbf{y}_i^{(t)+k} ; \xi^{(t)+k}\right)+\frac{K}{n} \sum_i \mathbf{c}_i^{(t)}\right\|^2 \nonumber\\
& \stackrel{(a)}{\leq} \frac{2 K \eta_x^2}{n} \sum_{i, k} \mathbb{E}\left\|\nabla_x f_i\left(\mathbf{x}_i^{(t)+k}, \mathbf{y}_i^{(t)+k}\right)-\nabla_x f_i\left(\bar{\mathbf{x}}_i^{(t)}, \bar{\mathbf{y}}_i^{(t)}\right)\right\|^2 +2 K^2 \eta_x^2 \mathbb{E}\left\|\nabla_x f\left(\bar{\mathbf{x}}_i^{(t)}, \bar{\mathbf{y}}_i^{(t)}\right)\right\|^2\nonumber\\&\quad+\frac{K \eta^2_x \sigma^2}{n} \nonumber\\
& \leq \frac{2 K \eta_x^2 \ell^2}{n} \sum_{i, k}\left(\mathbb{E}\left\|\mathbf{x}_i^{(t)+k}-\bar{\mathbf{x}}^{(t)}\right\|^2+\mathbb{E}\left\|\mathbf{y}_i{ }^{(t)+k}-\bar{\mathbf{y}}^{(t)}\right\|^2\right)+2 K^2 \eta_x^2 \mathbb{E}\left\|\nabla_x f\left(\bar{\mathbf{x}}_i{ }^{(t)}, \bar{\mathbf{y}}_i^{(t)}\right)\right\|^2\nonumber\\&\quad+\frac{K \eta^2_x \sigma^2}{n} \nonumber\\
& \stackrel{(b)}{\leq} 2 K \eta_x^2 \ell^2\left(\mathcal{E}_t^x+\mathcal{E}_t^y\right)+2 K^2 \eta_x^2\left(2 \ell^2 \delta_t+2 \mathbb{E}\left\|\nabla \Phi\left(\bar{\mathbf{x}}^{(t)}\right)\right\|^2\right)+\frac{K \eta^2_x \sigma^2}{n}.\label{eq.4}
\end{align}
Similar to the above derivations, we have
\begin{align}\label{eq.5}
\Delta_{t+1}^y :=\mathbb{E}\left\|\bar{\mathbf{y}}^{(t+1)}-\bar{\mathbf{y}}^{(t)}\right\|^2 &\leq 2 K^2 \eta_y^2 \ell^2\left(\mathcal{E}_t^x+\mathcal{E}_t^y\right)+2 K^2 \eta_y^2 \mathbb{E}\left\|\nabla_y f\left(\bar{\mathbf{x}}_i^{(t)}, \bar{\mathbf{y}}_i^{(t)}\right)\right\|^2+\frac{K \eta_y^2 \sigma^2}{n}\nonumber\\
& \stackrel{(c)}{\leq} 2 K \eta_y^2 \ell^2\left(\mathcal{E}_t^x+\mathcal{E}_t^y\right)+2 K^2 \eta_y^2 \ell^2 \delta_t+\frac{K \eta_y^2 \sigma^2}{n}.
\end{align}
We used Lemma \ref{lem.3}, \ref{lem.4}, and \ref{lem.4} in $ (a) $, $ (b) $, and $ (c) $, respectively. Combining \eqref{eq.4} and \eqref{eq.5} completes the proof.
\hfill \ensuremath{\Box} 
\begin{lemmap}{A.8}\label{lem.8}
Assuming that $ \eta_x, \eta_y \leq \frac{\sqrt{p}}{\sqrt{24} K \ell} $, we have the following bounds on the quality of correction for variables $ \mathbf{x} $ and $ \mathbf{y} $
\begin{align}\label{eq.6}
 \gamma_{t+1}^x &\leq\left(1-\frac{p}{2}\right) \gamma_t^x+\frac{25}{pK}\left(\mathcal{E}_t^x+\mathcal{E}_t^y\right)+\frac{12 K^2 \ell^2}{p}\left(2 \eta_x^2+\eta_y^2\right) \delta_t +\frac{24 K^2 \eta_x^2}{p}\mathbb{E}\left\|\nabla \Phi\left(\bar{\mathbf{x}}^{(t)}\right)\right\|^2+\frac{2\sigma^2}{K \ell^2},\\
\label{eq.7}
 \gamma_{t+1}^y &\leq\left(1-\frac{p}{2}\right) \gamma_t^y+\frac{25}{pK}\left(\mathcal{E}_t^x+\mathcal{E}_t^y\right)+\frac{12 K^2 \ell^2}{p}\left(2 \eta_x^2+\eta_y^2\right) \delta_t +\frac{24 K^2 \eta_x^2}{p}\mathbb{E}\left\|\nabla \Phi\left(\bar{\mathbf{x}}^{(t)}\right)\right\|^2+\frac{2\sigma^2}{K \ell^2}.
\end{align}
\end{lemmap}
\textit{Proof.} We can write that
\begin{equation}
\begin{aligned}
 n \ell^2 \gamma_{t+1}^x &:=\mathbb{E}\left\|\mathbf{C}^{(t+1)}+\nabla_x f\left(\bar{\mathbf{X}}^{(t+1)}, \bar{\mathbf{Y}}^{(t+1)}\right)(\mathbf{I}-\mathbf{J})\right\|_F^2 \\
& =\mathbb{E} \left\| \mathbf{C}^{(t)} \mathbf{W}+\frac{1}{K} \sum_{k=0}^{K-1} \nabla_x F\left(\mathbf{X}^{(t)+k}, \mathbf{Y}^{(t)+k} ; \xi^{(t)+k}\right)(\mathbf{W}-\mathbf{I}) +\nabla_x f\left(\bar{\mathbf{X}}^{(t+1)}, \bar{\mathbf{Y}}^{(t+1)}\right)(\mathbf{I}-\mathbf{J}) \right\|_F^2 \\
& \leq \mathbb{E} \left\|\left(\mathbf{C}^{(t)}+\nabla_x f\left(\bar{\mathbf{X}}^{(t)}, \bar{\mathbf{Y}}^{(t)}\right)(\mathbf{I}-\mathbf{J})\right) \mathbf{W} \right.\\
& \left. \quad+\left(\frac{1}{K} \sum_{k=0}^{K-1} \nabla_x f\left(\mathbf{X}^{(t)+k}, \mathbf{Y}^{(t)+k}\right)-\nabla_x f\left(\bar{\mathbf{X}}^{(t)}, \bar{\mathbf{Y}}^{(t)}\right)\right)(\mathbf{W}-\mathbf{I}) \right.\\
& \left. \quad+\left(\nabla_x f\left(\bar{\mathbf{X}}^{(t+1)}, \bar{\mathbf{Y}}^{(t+1)}\right)-\nabla_x f\left(\bar{\mathbf{X}}^{(t)}, \bar{\mathbf{Y}}^{(t)}\right)\right)(\mathbf{I}-\mathbf{J}) \right\|_F^2+\frac{n \sigma^2}{K} \\
& \stackrel{(a)}{\leq}(1+\alpha)(1-p) n \ell^2 \gamma_t^x \\
& \quad +2\left(1+\frac{1}{\alpha}\right)\left(\|\mathbf{W}-\mathbf{I}\|^2 \frac{\ell^2}{K} \sum_{k=0}^{K-1}\left(\mathbb{E}\left\|\mathbf{X}^{(t)+k}-\bar{\mathbf{X}}^{(t)}\right\|^2+\mathbb{E}\left\|\mathbf{Y}^{(t)+k}-\bar{\mathbf{Y}}^{(t)}\right\|^2\right)\right. \\
& \quad \left. +\|\mathbf{I}-\mathbf{J}\|^2 n \ell^2\left(\mathbb{E}\left\|\bar{\mathbf{x}}^{(t+1)}-\bar{\mathbf{x}}^{(t)}\right\|^2+\mathbb{E}\left\|\bar{\mathbf{y}}^{(t+1)}-\bar{\mathbf{y}}^{(t)}\right\|^2\right)\right)+\frac{n \sigma^2}{K} \\
& \stackrel{\alpha = \frac{p}{2}, \frac{1}{p}\geq 1}{\leq}\left(1-\frac{p}{2}\right) n \ell^2 \gamma_t^x+\frac{6}{p}\left(\frac{4 \ell^2 n}{K}\left(\mathcal{E}_t^x+\mathcal{E}_t^y\right)+n \ell^2\left(\Delta_{t+1}^x+\Delta_{t+1}^y\right)\right)+\frac{n \sigma^2}{K}.\nonumber
\end{aligned}
\end{equation}
In $ (a) $ we applied Assumption \ref{assm: W} and the fact that
  \begin{equation}
  \begin{aligned}
     &\left(\mathbf{C}^{(t)}+\nabla_x f\left(\bar{\mathbf{X}}^{(t)}, \bar{\mathbf{Y}}^{(t)}\right)(\mathbf{I}-\mathbf{J})\right)\mathbf{J} = \mathbf{C}^{(t)}\mathbf{J} + \nabla_x f\left(\bar{\mathbf{X}}^{(t)}, \bar{\mathbf{Y}}^{(t)}\right)(\mathbf{J}-\mathbf{J}) 
     \stackrel{\text{Lemma \ref{lem.3}}}{=} \mathbf{0}. \nonumber
  \end{aligned}
\end{equation}  
Using Lemma \ref{lem.7} to bound $ \Delta_{t+1}^x+\Delta_{t+1}^y $ we have 
\begin{equation}
\begin{aligned}
 \gamma_{t+1}^x &\leq\left(1-\frac{p}{2}\right) \gamma_t^x+\frac{1}{p}\left(\frac{24}{K}+12 K \eta_x^2 \ell^2+12 K \eta_y^2 \ell^2\right)\left(\mathcal{E}_t^x+\mathcal{E}_t^y\right)+\frac{12 K^2 \ell^2}{p}\left(2 \eta_x^2+\eta_y^2\right) \delta_t \\
&\quad +\frac{24 K^2 \eta_x^2}{p}\mathbb{E}\left\|\nabla \Phi\left(\bar{\mathbf{x}}^{(t)}\right)\right\|^2+\frac{6 K \sigma^2\left(\eta_x^2+\eta_y^2\right)}{n p}+\frac{\sigma^2}{K \ell^2}.\nonumber
\end{aligned}
\end{equation}
Applying the conditions on the step sizes will result in \eqref{eq.6}. In a similar fashion, we can show \eqref{eq.7}.
\hfill \ensuremath{\Box} 
\begin{lemmap}{A.9}\label{lem.9}
Using Proposition \ref{prop2} and assuming that $ \eta_y \leq \frac{1}{K\ell} $, we have the following bound on \\$\mathbb{E}\left\|\hat{\mathbf{y}}^{(t)}-\bar{\mathbf{y}}^{(t+1)}\right\|^2$ for any $ \alpha > 0$:
\begin{equation}
\begin{aligned}
    \mathbb{E}\left\|\hat{\mathbf{y}}^{(t)}-\bar{\mathbf{y}}^{(t+1)}\right\|^2 &\leq(1+\alpha)\left(1-K \eta_y \mu\right) \delta_t+\left(1+\frac{1}{\alpha}\right) \eta_y^2 \ell^2 K\left(\mathcal{E}^x+\mathcal{E}^y\right) +\frac{K \eta^2_y \sigma^2}{n}.\nonumber
\end{aligned}
\end{equation}
\end{lemmap}
\textit{Proof.}
If we replace $ \mathbf{x} = \bar{\mathbf{x}}^{(t)} $, $ \mathbf{y} = \bar{\mathbf{y}}^{(t)} $, and $ \mathbf{y}' = \hat{\mathbf{y}}^{(t)} $ in Proposition \ref{prop2}, we have
\begin{align}
    \nabla_y f(\bar{\mathbf{x}}^{(t)}, \bar{\mathbf{y}}^{(t)})^{\top}&(\bar{\mathbf{y}}^{(t)}-\hat{\mathbf{y}}^{(t)})+\frac{1}{2 \ell}\left\|\nabla_y f(\bar{\mathbf{x}}^{(t)}, \bar{\mathbf{y}}^{(t)})\right\|^2+\frac{\mu}{2}\|\bar{\mathbf{y}}^{(t)}-\hat{\mathbf{y}}^{(t)}\|^2 \leq 0.\label{eq.8}
\end{align}
We can also write that
\begin{equation}
    \begin{aligned}
 \mathbb{E}&\left\|\hat{\mathbf{y}}^{(t)}-\bar{\mathbf{y}}^{(t)}-K \eta_y \nabla_y f\left(\bar{\mathbf{x}}^{(t)}, \bar{\mathbf{y}}^{(t)}\right)\right\|^2  \\
 &=\mathbb{E}\left\|\hat{\mathbf{y}}^{(t)}-\bar{\mathbf{y}}^{(t)}\right\|^2-2 K \eta y \mathbb{E}\left\langle\hat{\mathbf{y}}^{(t)}-\bar{\mathbf{y}}^{(t)}, \nabla y f\left(\bar{\mathbf{x}}^{(t)}, \bar{\mathbf{y}}^{(t)}\right)\right\rangle +K^2 \eta_y^2 \mathbb{E}\left\|\nabla_y f\left(\bar{\mathbf{x}}^{(t)}, \bar{\mathbf{y}}^{(t)}\right)\right\|^2 \\
& =\mathbb{E}\left\|\hat{\mathbf{y}}^{(t)}-\bar{\mathbf{y}}^{(t)}\right\|^2+2 K \eta_y\left(\mathbb{E}\left\langle\bar{\mathbf{y}}^{(t)}-\hat{\mathbf{y}}(t), \nabla_y f\left(\bar{\mathbf{x}}^{(t)}, \bar{\mathbf{y}}^{(t)}\right)\right\rangle +\frac{K \eta_y}{2} \mathbb{E}\left\|\nabla_y f\left(\bar{\mathbf{x}}^{(t)}, \bar{\mathbf{y}}^{(t)}\right)\right\|^2\right) \\
& \stackrel{(a)}{\leq} \mathbb{E}\left\|\hat{\mathbf{y}}^{(t)}-\bar{\mathbf{y}}^{(t)}\right\|^2+2 K \eta_y\left(-\frac{\mu}{2} \mathbb{E}\left\|\hat{\mathbf{y}}^{(t)}-\bar{\mathbf{y}}^{(t)}\right\|^2\right) =\left(1-K \eta_y \mu\right) \delta_t.\nonumber
\end{aligned}
\end{equation}
In $ (a) $, we used the assumption that $ \eta_y \leq \frac{1}{K\ell} $ and \eqref{eq.8}. Now, we can write 
\begin{equation}
\begin{aligned}
 \mathbb{E}\left\|\hat{\mathbf{y}}^{(t)}-\bar{\mathbf{y}}^{(t+1)}\right\|^2 & \stackrel{(b)}{=} \mathbb{E}\left\|\hat{\mathbf{y}}^{(t)}-\bar{\mathbf{y}}^{(t)}-\frac{\eta_y}{n} \sum_{i, k} \nabla_y F_i\left(\mathbf{x}_i^{(t)+k}, \mathbf{y}_i^{(t)+k} ; \xi^{(t)+k}\right)\right\|^2 \\
& \leq \mathbb{E} \left\| \hat{\mathbf{y}}^{(t)}-\bar{\mathbf{y}}^{(t)}-K \eta_y \nabla_y f\left(\bar{\mathbf{x}}^{(t)}, \bar{\mathbf{y}}^{(t)}\right)-\frac{\eta_y}{n} \sum_{i, k} \nabla_y f_i\left(\mathbf{x}_i^{(t)+k}, \mathbf{y}_i^{(t)+k}\right)\right. \\
& \quad \left. +\frac{\eta_y}{n} \sum_{i, k} \nabla_y f_i\left(\bar{\mathbf{x}}^{(t)}, \bar{\mathbf{y}}^{(t)}\right)\right \|^2+\frac{K \eta^2_y \sigma^2}{n} \\
& \leq(1+\alpha) \mathbb{E}\left\|\hat{\mathbf{y}}^{(t)}-\bar{\mathbf{y}}^{(t)}-K \eta_y \nabla_y f\left(\bar{\mathbf{x}}^{(t)}, \bar{\mathbf{y}}^{(t)}\right)\right\|^2 + \frac{K \eta^2_y \sigma^2}{n} \\
& \quad+\left(1+\frac{1}{\alpha}\right) \frac{\eta_y^2 K}{n} \sum_{i, k} \mathbb{E}\left\|\nabla_y f_i\left(\mathbf{x}_i^{(t)+k}, y_i^{(t)+k}\right)-\nabla_y f_i\left(\bar{\mathbf{x}}^{(t)}, \bar{\mathbf{y}}^{(t)}\right)\right\|^2 \\
& \leq(1+\alpha)\left(1-K \eta_y \mu\right) \delta_t+\left(1+\frac{1}{\alpha}\right) \eta_y^2 \ell^2 K\left(\mathcal{E}^x+\mathcal{E}^y\right) +\frac{K \eta^2_y \sigma^2}{n}.\nonumber
\end{aligned}
\end{equation}
where in $ (b) $, we used Lemma \ref{lem.3}; i.e., $ \frac{1}{n}\sum_i \mathbf{d}_i^{(t)} = \mathbf{0} $.
\hfill \ensuremath{\Box} 
\begin{lemmap}{A.10}\label{lem.10}
Assuming that $ \eta_x \leq \frac{ \eta_y}{4 \sqrt{6} \kappa^2} $ and $ \eta_y \leq \frac{1}{K\ell} $, we have the following bound on $ \delta_t $ 
\begin{equation}
\begin{aligned}
 \delta_{t+1} &\leq \left(1-\frac{K \eta_y \ell}{6\kappa}\right) \delta_t+12 \eta_y \ell\kappa \left(\mathcal{E}_t^x+\mathcal{E}_t^y\right)+\frac{16 \kappa^3 K \eta_x^2}{\eta_y \ell}\mathbb{E}\left\|\nabla \Phi\left(\bar{\mathbf{x}}^{(t)}\right)\right\|^2 +\frac{8 \eta_y \sigma^2 \kappa}{n \ell}.   \nonumber
\end{aligned}
\end{equation}
\end{lemmap}
\textit{Proof.} We begin the proof by writing that
\begin{equation}
\begin{aligned}
 \delta_{t+1} &\stackrel{(a)}{\leq}(1+\beta) \mathbb{E}\left\|\hat{\mathbf{y}}^{(t)}-\bar{\mathbf{y}}^{(t+1)}\right\|^2+\left(1+\frac{1}{\beta}\right) \mathbb{E}\left\|\hat{\mathbf{y}}^{(t+1)}-\hat{\mathbf{y}}^{(t)}\right\|^2 \\
& \leq(1+\beta)(1+\alpha)\left(1-K \eta_y \mu\right) \delta_t+(1+\beta)\left(1+\frac{1}{\alpha}\right) \eta_y^2 \ell^2 K\left(\mathcal{E}_t^x+\mathcal{E}_t^y\right) \\
& \quad +(1+\frac{1}{\beta}) \kappa^2 \mathbb{E}\left\|\bar{\mathbf{x}}^{(t+1)}-\bar{\mathbf{x}}^{(t)}\right\|^2+(1+\beta) \frac{K \eta_y^2 \sigma^2}{n} \\
& \stackrel{(b)}{\leq}\left(1-\frac{K \eta_y \mu}{3}\right) \delta_t+\frac{6 \eta_y \ell^2}{\mu}\left(\mathcal{E}_t^x+\mathcal{E}_t^y\right)+\frac{4 \eta_y \sigma^2}{n \mu} \\
&\quad+\frac{4 \kappa^2}{K \eta_y \mu}\left(2 K \eta_x^2 \ell^2\left(\mathcal{E}_t^x+\mathcal{E}_t^y\right)+4 K^2 \ell^2 \eta_x^2 \delta_t+4 K^2 \eta_x^2 \mathbb{E}\|\nabla \Phi(\bar{\mathbf{x}}(t))\|^2+\frac{K \eta_x^2 \sigma^2}{n}\right) \\
&=\left(1-\frac{K \eta_y \ell}{3\kappa}+\frac{16\ell\kappa^3 K \eta_x^2}{\eta_y}\right) \delta_t+\left(\frac{8 \ell \kappa^3 \eta_x^2}{\eta_y}+6 \eta_y \ell\kappa\right)\left(\mathcal{E}_t^x+\mathcal{E}_t^y\right)+\frac{16 \kappa^3 K \eta_x^2}{\eta_y \ell}\mathbb{E}\left\|\nabla \Phi\left(\bar{\mathbf{x}}^{(t)}\right)\right\|^2 \\
&\quad+\frac{4 \kappa^3 \eta_x^2 \sigma^2}{n \eta_y \ell}+\frac{4 \eta_y \sigma^2\kappa}{n \ell}.\nonumber
\end{aligned}
\end{equation}
Using the assumption $ \eta_x \leq \frac{ \eta_y}{4 \sqrt{6} \kappa^2} $ completes the proof.
In $ (a) $, we used the bound in Lemma \ref{lem.9} for the first term and Proposition \ref{prop1} for the second term. In $ (b) $, we replaced $ \alpha = \beta = \frac{K\eta_y \mu}{3} $ and used \eqref{eq.4} in Lemma \ref{lem.7}.
\hfill \ensuremath{\Box} 
\begin{lemmap}{A.11}\label{lem.11}
Assuming that $ \eta_x \leq \frac{1}{16K\ell \kappa}$, we have the following bound on $ \mathbb{E} \Phi\left(\bar{\mathbf{x}}^{(t+1)}\right) $ as follows
\begin{equation}
\begin{aligned}
    \mathbb{E} \Phi\left(\bar{\mathbf{x}}^{(t+1)}\right)&\leq \mathbb{E} \Phi\left(\bar{\mathbf{x}}^{(t)}\right)+2 \eta_x \ell^2\left(\mathcal{E}_t^x+\mathcal{E}_t^y\right)+2\ell^2 \eta_x K \delta_t-\frac{\eta_x K}{4} \mathbb{E}\left\|\nabla \Phi\left(\bar{\mathbf{x}}^{(t)}\right)\right\|^2+\frac{K \eta_x^2 \ell \sigma^2\kappa}{n}.\nonumber
\end{aligned}
\end{equation}
\end{lemmap}
\textit{Proof.}
According to the Proposition \ref{prop1}, $ \Phi(\cdot) $ is $2\kappa\ell$-smooth, which results in the following
\begin{equation}
\begin{aligned}
 \mathbb{E} \Phi\left(\bar{\mathbf{x}}^{(t+1)}\right)&= \mathbb{E} \Phi\left(\bar{\mathbf{x}}^{(t)}-\frac{\eta_x}{n} \sum_{i, k}\left(\nabla_x F_i\left(\mathbf{x}_i^{(t)+k}, \mathbf{y}_i^{(t)+k} ; \xi_i^{(t)+k}\right)+\mathbf{c}_i^{(t)}\right)\right) \\
& \leq \mathbb{E} \Phi\left(\bar{\mathbf{x}}^{(t)}\right)+\underbrace{\mathbb{E}\left\langle\nabla \Phi\left(\bar{\mathbf{x}}^{(t)}\right), \frac{-\eta_x}{n} \sum_{i, k}\left(\nabla_x F_i\left(\mathbf{x}_i^{(t)+k}, \mathbf{y}_i^{(t)+k} ; \xi_i^{(t)+k}\right)+\mathbf{c}_i^{(t)}\right)\right\rangle}_{:=U} \\
&\quad +\kappa \ell \mathbb{E}\left\|\bar{\mathbf{x}}^{(t+1)}-\bar{\mathbf{x}}^{(t)}\right\|^2.\nonumber
\end{aligned}
\end{equation}
Now, we derive an upper bound for $ U $ as follows
\begin{equation}
\begin{aligned}
 U&:=\mathbb{E}\left\langle \nabla \Phi\left(\bar{\mathbf{x}}^{(t)}\right),-\frac{\eta_x}{n} \sum_{i, k}\left(\nabla_x F_i\left(\mathbf{x}_i^{(t)+k}, \mathbf{y}_i^{(t)+k} ; \xi_i^{(t)+k}\right)+\mathbf{c}_i^{(t)}\right)\right\rangle \\
& =\mathbb{E}\left\langle\nabla \Phi\left(\bar{\mathbf{x}}^{(t)}\right),-\frac{\eta_x}{n} \sum_{i, k} \mathbb{E}_{\xi_i^{(t)+k}} \nabla_x F_i\left(\mathbf{x}_i^{(t)+k}, \mathbf{y}_i^{(t)+k} ; \xi_i^{(t)+k}\right)\right\rangle \\
& =-\eta_x \mathbb{E}\left\langle\nabla \Phi\left(\bar{\mathbf{x}}^{(t)}\right), \frac{1}{n} \sum_{i, k}\left(\nabla_x f_i\left(\mathbf{x}_i^{(t)+k}, \mathbf{y}_i^{(t)+k}\right)-\nabla_x f_i\left(\bar{\mathbf{x}}^{(t)}, \bar{\mathbf{y}}^{(t)}\right)\right.\right. \\
& \left.\left.\quad+\nabla_x f_i\left(\bar{\mathbf{x}}^{(t)}, \bar{\mathbf{y}}^{(t)}\right)-\nabla_x f_i\left(\bar{\mathbf{x}}^{(t)}, \hat{\mathbf{y}}^{(t)}\right)+\nabla_x f_i\left(\bar{\mathbf{x}}^{(t)}, \hat{\mathbf{y}}^{(t)}\right)\right)\right\rangle \\
& =-K \eta_x \mathbb{E}\left\|\nabla \Phi\left(\bar{\mathbf{x}}^{(t)}\right)\right\|^2-\frac{\eta_x}{n} \sum_{i, k}\left\langle\nabla \Phi\left(\bar{\mathbf{x}}^{(t)}\right), \nabla_x f_i\left(\mathbf{x}_i^{(t)+k}, \mathbf{y}_i^{(t)+k}\right)-\nabla_x f_i\left(\bar{\mathbf{x}}^{(t)}, \bar{\mathbf{y}}^{(t)}\right)\right. \\
&\left. \quad +\nabla_x f_i\left(\bar{\mathbf{x}}^{(t)}, \bar{\mathbf{y}}^{(t)}\right)-\nabla_x f_i\left(\bar{\mathbf{x}}^{(t)}, \hat{\mathbf{y}}^{(t)}\right)\right\rangle \\
& \leq-\frac{K \eta_x}{2} \mathbb{E}\left\|\nabla \Phi\left(\bar{\mathbf{x}}^{(t)}\right)\right\|^2+\frac{\eta_{x}}{n} \sum_{i, k}\left(\mathbb{E}\left\|\nabla_x f_i\left(\mathbf{x}_i^{(t)+k}, \mathbf{y}_i^{(t)+k}\right)-\nabla_x f_i\left(\bar{\mathbf{x}}^{(t)}, \bar{\mathbf{y}}^{(t)}\right)\right\|^2\right. \\
& \quad\left.+\mathbb{E}\left\|\nabla_x f_i\left(\bar{\mathbf{x}}^{(t)}, \bar{\mathbf{y}}^{(t)}\right)-\nabla_x f_i\left(\bar{\mathbf{x}}^{(t)}, \hat{\mathbf{y}}^{(t)}\right)\right\|^2\right) \\
& \leq-\frac{K \eta_x}{2} \mathbb{E}\left\|\nabla \Phi\left(\bar{\mathbf{x}}^{(t)}\right)\right\|^2+\eta_x \ell^2\left(\mathcal{E}_t^x+\mathcal{E}_t^y\right)+K \eta_x \ell^2 \delta_t.\nonumber
\end{aligned}
\end{equation}
Now, we apply the above upper bound for $ U $ and \eqref{eq.4} in Lemma \ref{lem.7} as follows
\begin{equation}
\begin{aligned}
 \mathbb{E} \Phi\left(\bar{\mathbf{x}}^{(t+1)}\right) &\leq \mathbb{E} \Phi\left(\bar{\mathbf{x}}^{(t)}\right)+\eta_x \ell^2\left(\mathcal{E}_t^x+\mathcal{E}_t^y\right)+ \ell^2 \eta_x K \delta_t-\frac{\eta_x K}{2} \mathbb{E}\left\|\nabla \phi\left(\bar{\mathbf{x}}^{(t)}\right)\right\|^2 +\kappa \ell \mathbb{E}\left\|\bar{\mathbf{x}}^{(t+1)}-\bar{\mathbf{x}}^{(t)}\right\|^2 \\
& \leq \mathbb{E} \Phi\left(\bar{\mathbf{x}}^{(t)}\right)+\left(\eta_x \ell^2+2 K \eta_x^2 \ell^3\kappa\right)\left(\mathcal{E}_t^x+\mathcal{E}_t^y\right)+\frac{K \eta_x^2 \ell\kappa \sigma^2}{n} \\
&\quad+\left( \ell^2 \eta_x K+4 K^2 \ell^3 \eta_x^2\kappa\right) \delta_t+\left(4 K^2 \eta_x^2 \ell\kappa-\frac{\eta_x K}{2}\right) \mathbb{E}\left\|\nabla \phi\left(\bar{\mathbf{x}}^{(t)}\right)\right\|^2.\nonumber
\end{aligned}
\end{equation}
Applying the assumption $ \eta_x \leq \frac{1}{16K\ell \kappa}$ completes the proof.
\hfill \ensuremath{\Box} 
\begin{lemmap}{A.12}\label{lem.12}
Under the assumption that $ \eta_d = \Theta(\frac{p}{\kappa K\ell}) $, $ \eta_c = \Theta(\frac{\eta_d}{\kappa^2}) $, and $ \eta_s = \eta_r = \Theta(p) $, we can find constants $ A_x $, $ A_y $, $ B_x $, $ B_y $, and $ C $, such that $ D>0 $ and $ D_9 \geq 0 $, and we have
\begin{equation}\label{eq.9}
\begin{aligned}
\mathcal{H}_{t+1}-\mathcal{H}_t &\leq-D K \eta_x \mathbb{E}\left\|\nabla \Phi\left(\bar{\mathbf{x}}^{(t)}\right)\right\|^2+D_9 K\ell \eta_d^3 \sigma^2+\frac{K \eta_x^2 \ell \kappa}{n} \sigma^2+\frac{8 \eta_y}{n  p} \sigma^2,
\end{aligned}
\end{equation}
where
\begin{equation}
\begin{aligned}
   \mathcal{H}_t&=\mathbb{E} \Phi\left(\bar{\mathbf{x}}^{(t)}\right)-\mathbb{E} \Phi\left(\mathbf{x}^*\right)+A_x \eta_d K \ell^2 \Xi_t^x+A_y \eta_d K \ell^2 \Xi_t^y+B_x K^3 \ell^4 \eta_d^3 \gamma_t^x+B_y K^3 \ell^4 \eta_d^3 \gamma_t^y 
+C \frac{\ell}{ \kappa p} \delta_t. \nonumber
\end{aligned}
\end{equation}
\end{lemmap}
\textit{Proof.}
According to the Lemma \ref{lem.5}, we have
\begin{equation}
\begin{aligned}
& 0 \leq-D_x \ell^2 \eta_d \mathcal{E}_t^x+3 D_x K \ell^2 \eta_d \Xi_t^x+12 D_x K^2 \eta_c^2 \eta_d \ell^4 \mathcal{E}_t^y+12 D_x K^3 \eta_c^2 \eta_d \ell^4 \gamma_t^x+12 D_x K^3 \eta_c^2 \eta_d \ell^4 \delta_t 
\\
&\quad+12 D_x K^3 \eta_c^2 \eta_d \ell^2 \mathbb{E}\left\|\nabla \Phi\left(\bar{\mathbf{x}}^{(t)}\right)\right\|^2+3 D_x K^2 \eta_c^2 \eta_d \ell^2 \sigma^2, \\
& 0 \leq-D_y \ell^2 \eta_d \mathcal{E}_t^y+3 D_y K \ell^2 \eta_d \Xi_t^y+12 D_y K^2 \eta_d^3 \ell^4  \mathcal{E}_t^x+12 D_y K^3 \eta_d^3 \ell^4 \gamma_t^y+6 D_y K^3 \eta_d^3 \ell^4 \delta_t+3 D_y K^2 \eta_d^3 \ell^2 \sigma^2.\label{eq.10}
\end{aligned}
\end{equation}
By applying the definition of $ \mathcal{H}_t $ from \eqref{eq.9} and using \eqref{eq.10}, Lemmas~\ref{lem.5}, \ref{lem.6}, \ref{lem.8}, \ref{lem.10}, and \ref{lem.11}, we have
\begin{footnotesize}
\begin{align}
& \mathcal{H}_{t+1}-\mathcal{H}_t \leq\underbrace{\left(-B_x \frac{p}{2}+A_x \frac{6 \eta_s^2}{p}+D_x 12\right)}_{\leq D_1} \eta_d^3 K^3 \ell^4 \gamma_t^x \nonumber\\
& +\underbrace{\left(-B_y \frac{p}{2}+A_y \frac{6 \eta_r^2}{p}+D_y 12\right)}_{\leq D_2} \eta_d^3 K^3 \ell^4 \gamma_t^y \nonumber\\
&+\underbrace{\left(-A_x \frac{p}{2}+3 D_x\right)}_{\leq D_3} \Xi_t^x \eta_d K \ell^2 \nonumber\\
& +\underbrace{\left(-A_y \frac{p}{2}+3 D_y\right)}_{\leq D_4} \Xi_t^y \eta_d K \ell^2 \nonumber\\
& +\underbrace{\left(-D_x+A_x \frac{6 K^2 \ell^2 \eta_x^2}{p}+A_y \frac{6 K^2 \ell^2 \eta_y^2}{p}+B_x \frac{25 \eta_d^2 \ell^2 K^2}{p}+B_y \frac{25 \eta_d^2 \ell^2 K^2}{p}+D_y 12 K^2 \ell^2 \eta_d^2+\frac{2 \eta_x }{\eta_d}+C \frac{12 \eta_r}{p}\right)}_{\leq D_5} \ell^2 \eta_d \mathcal{E}_t^x \nonumber\\
& +\underbrace{\left(-D_y+A_x \frac{6 K^2 \ell^2 \eta_x^2}{p}+A_y \frac{6 K^2 \ell^2 \eta_y^2}{p}+B_x \frac{25 \eta_d^2 \ell^2 K^2}{p}+B_y \frac{25 \eta_d^2 \ell^2 K^2}{p}+D_x 12 K^2 \ell^2 \eta_c^2+\frac{2 \eta_x }{\eta_d}+C \frac{12 \eta_r}{p}\right)}_{\leq D_6} \ell^2 \eta_d \mathcal{E}_t^y \nonumber\\
&+\underbrace{\left(-C \frac{\eta_r}{6p}+B_x \frac{12 K^4 \ell^4}{p} \eta_d^2\left(3\eta_y^2\right)\kappa^2+B_y \frac{12 K^4 \ell^4}{p} \eta_d^2\left(3\eta_y^2\right)\kappa^2+D_x 12 K^2 \ell^2 \eta_c^2\kappa^2+D_y 6 K^2 \ell^2 \eta_d^2\kappa^2+2  \kappa^2 \frac{\eta_x}{\eta_d}\right)}_{\leq D_7} \frac{K \ell^2 \eta_d}{\kappa^2} \delta_t\nonumber\\
& +\underbrace{\left(-\frac{1}{4}+B_x \frac{24 K^4 \ell^4}{p} \eta_d^3 \eta_x+B_y \frac{24 K^4 \ell^4}{p} \eta_d^3 \eta_x+C \frac{16 \kappa^2 \eta_x}{ \eta_y p}+D_x 12 K^2 \ell^2 \eta_d \frac{\eta_c}{\eta_s}\right)}_{\leq D_8} K \eta_x \mathbb{E}\left\|\nabla \Phi\left(\bar{\mathbf{x}}^{(t)}\right)\right\|^2 \nonumber\\
& +\underbrace{\left(A_x \eta_s^2+A_y \eta_r^2+B_x 2+B_y 2+D_x 3+D_y 3\right)}_{\leq D_9} K^2 \ell^2 \eta_d^3 \sigma^2+\frac{K \eta_x^2 \ell \kappa}{n} \sigma^2+C \frac{8 \eta_y}{n  p} \sigma^2.
&\nonumber
\end{align}
\end{footnotesize}

Assuming that $ D_x = D_y = v $, as long as $ \eta_d \leq \frac{p}{200v\kappa K\ell} $, $ \eta_c \leq \frac{\eta_d}{\kappa^2} $, $ \eta_s = \eta_r = pv $, $ A_x = A_y = \frac{6v}{p} $, $ B_x = B_y = \frac{1}{p}(72v^3 + 24v) $, and $ C = \frac{1}{24} $, there exists $ v > 1 $ that makes $ D_1 $, $ D_2 $, $ D_3 $, $ D_4 $, $ D_5 $, $ D_6 $, $ D_7 $ $ \leq 0 $, $ D_8 \leq -D < 0 $, and $ D_9 \geq 0 $.
\hfill \ensuremath{\Box} 
\\~\\
\textbf{Theorem A.1.}
Suppose  Assumptions \ref{assm: W}-\ref{assm: lower bound} hold and consider the iterates of Dec-FedTrack in Algorithm \ref{al.1} with step-sizes  $\eta_d=$ $\Theta\left(\frac{p}{\kappa K \ell}\right), \eta_c=\Theta\left(\frac{\eta_d}{\kappa^2}\right)$, and $\eta_s=\eta_r=\Theta(p)$. Then, after $T$ communication rounds each with $K$ local updates, there exists an iterate $0 \leq t \leq T$ such that $\mathbb{E}\|\nabla \Phi(\bar{\mathbf{x}}^{(t)})\|^2 \leq \epsilon^2$ for
\begin{align}
T=O\left(\frac{\kappa^3}{p^2 \epsilon^2}\right) \mathcal{H}_0 \ell, \quad
K = O\left(\frac{p^2\sigma^2}{\kappa^2 n \epsilon^2} + \frac{\sigma^2}{\epsilon^2} + \frac{\kappa^2 \sigma^2}{n p \epsilon^2}\right),\nonumber
\end{align}
where  $\mathcal{H}_0=O\left(1+\frac{\ell \delta_0}{\kappa p}\right)$ and $\delta_0 = O\left(\frac{q}{\mu^2}\right)$.
\\~\\
\textit{Proof.}
Using the telescopic sum for $ \mathcal{H}_t $, we have
\begin{equation}
\begin{aligned}
 \frac{1}{T+1} \sum_{t=0}^T\left(\mathcal{H}_{t+1}-\mathcal{H}_t\right)&=\frac{1}{T+1}\left(\mathcal{H}_{T+1}-\mathcal{H}_0\right) \\
& \leq-D K \eta_x \frac{1}{T+1} \sum_{t=0}^T \mathbb{E}\left\|\nabla \Phi\left(\bar{\mathbf{x}}^{(t)}\right)\right\|^2+D_9 K^2 \ell^2 \eta_d^3 \sigma^2+\frac{K \eta_x^2 \ell \kappa}{n} \sigma^2+\frac{8 \eta_y}{n p} \sigma^2,\nonumber
\end{aligned}
\end{equation}
which results in 
\begin{align}
 &\frac{1}{T+1} \sum_{t=0}^T \mathbb{E}\left\|\nabla \Phi\left(\bar{\mathbf{x}}^{(t)}\right)\right\|^2 \leq \frac{\mathcal{H}_0-\mathcal{H}_{T+1}}{(T+1) D} \frac{1}{K \eta_x}+\frac{D_9 K \ell^2 \eta_d^3}{D \eta_x} \sigma^2+\frac{\eta_x \ell \kappa}{n D} \sigma^2+\frac{8 \eta_y}{n  D K p \eta_x}\sigma^2.   \label{eq.12} 
\end{align}
Now, we want to ensure $ \frac{1}{T+1} \sum_{t=0}^T \mathbb{E}\left\|\nabla \Phi\left(\bar{\mathbf{x}}^{(t)}\right)\right\|^2 \leq \epsilon^2 $ for any arbitrary $ \epsilon > 0 $, which is equivalent to bounding each term on the RHS of \eqref{eq.12} to the order of $ \epsilon^2 $. Given that $ D = \Theta(1) $, $ D_9 = O\left(\frac{1}{p}\right) $, $ \eta_x = \Theta\left(\frac{p^2}{\kappa^3 K \ell}\right) $, and $ \eta_y = \Theta\left(\frac{p^2}{\kappa K \ell}\right) $, we have
\begin{equation}
\begin{aligned}
    T &= O\left(\frac{\kappa^3}{p^2 \epsilon^2}\right)\mathcal{H}_0 \ell,\\
    K &= O\left(\frac{p^2\sigma^2}{\kappa^2 n \epsilon^2} + \frac{\sigma^2}{\epsilon^2} + \frac{\kappa^2 \sigma^2}{n p \epsilon^2}\right),\nonumber
\end{aligned}
\end{equation}
where $\mathcal{H}_0=O\left(1+\frac{\ell \delta_0}{\kappa p}\right)$ and $\delta_0 = O\left(\frac{q}{\mu^2}\right)$.
\hfill \ensuremath{\Box} 


\subsection{Adversarial Attacks}\label{appendix2}
We provide  descriptions of the attacks used in the numerical results section.
\begin{enumerate}
        \item \textbf{FGSM}~\cite{goodfellow2014explaining}: This method is a single-step adversarial attack designed to create adversarial examples by slightly perturbing the input to maximize the loss of a neural network. The FGSM attack perturbs the input \( a \) in the direction of the gradient of the loss with respect to the input. 
This is achieved by computing the gradient of the loss function \( f(\mathbf{x}, a, b) \), where \( \mathbf{x} \) represents the model parameters, \( a \) is the input, and \( b \) is the true label. 
The adversarial example is then generated as:
\[
a' = a + \epsilon \cdot \text{sign}(\nabla_a f(\mathbf{x}, a, b)),
\]
where \( \epsilon \) controls the magnitude of the perturbation.
\item 
\textbf{PGD}~\cite{kurakin2016adversarial}: The Projected Gradient Descent method is an iterative extension of FGSM, providing a stronger adversarial attack by applying FGSM multiple times with smaller step sizes. The PGD attack iteratively refines the adversarial example by applying small perturbations to the input. Starting from an initial adversarial example \( a_0 \) (often set to the original input \( a \)), the method updates the adversarial input \( a_t \) at each iteration using the formula:
\[
a_{t+1} = \text{Proj}_{\mathcal{B}_\epsilon(a)} \left( a_t + \eta \cdot \text{sign}(\nabla_a f(\mathbf{x}, a_t, b)) \right),
\]
where \( \eta \) is the step size, and \( \text{Proj}_{\mathcal{B}_\epsilon(x)} \) ensures the perturbed input remains within the \( L_\infty \)-norm ball of radius \( \epsilon \) around the original input.
\item \textbf{UAP}:
Universal Adversarial Perturbation is a technique designed to craft a single perturbation vector $\mathbf{y}$ that, when added to any input, significantly degrades the performance of a model. Unlike input-specific adversarial perturbations (e.g., FGSM or PGD), UAPs are input-agnostic and aim to generalize across a wide range of inputs. We use the universal perturbation introduced in~\cite{shafahi2020universal}, where the authors employ Stochastic Projected Gradient Descent (SPGD) to generate UAP. Their algorithm computes the gradient of the loss function \( f(\mathbf{x}, a + \mathbf{y}, b) \) with respect to \( \mathbf{y} \) as:
\[
g = \nabla_\mathbf{y} f(\mathbf{x}, a + \mathbf{y}, b).
\]
Using SPGD, \( \mathbf{y} \) is updated as:
\[
\mathbf{y} \leftarrow \mathbf{y} + \eta \cdot g,
\]
where \( \eta \) is the learning rate. After each update, \( \mathbf{y} \) is projected back onto the constraint set \( \|\mathbf{y}\|_p \leq \delta \) using:
\[
\mathbf{y} \leftarrow \text{Proj}_{\|\mathbf{y}\|_p \leq \delta}(\mathbf{y}).
\]
This process is iterated until \( \mathbf{y} \) achieves the desired attack success rate across the dataset.
\end{enumerate}


\end{document}